\documentclass{article}
 \usepackage[preprint]{neurips_2026}

\usepackage[utf8]{inputenc} 
\usepackage[T1]{fontenc}    
\usepackage{hyperref}       
\usepackage{url}            
\usepackage{booktabs}       
\usepackage{amsfonts}       
\usepackage{nicefrac}       
\usepackage{microtype}      
\usepackage{xcolor}         

\usepackage{lipsum}
\usepackage{graphicx}
\usepackage{duckuments}
\usepackage{makecell}
\usepackage{multirow}  
\usepackage{pifont}

\usepackage{amsmath} 
\newcommand{\paragrapht}[1]{\noindent\textbf{#1}}
\usepackage{cleveref}
\usepackage{comment}
\usepackage{amssymb}
\usepackage{subcaption}
\usepackage{wrapfig}

\title{\ours: Autoregressive 4D Generation with Temporal Structured Latents}

\usepackage{xspace}

\newcommand{\ours}{{MORPHOS}\xspace}
\newcommand{\tslat}{T-SL\textsc{at}\xspace}
\newcommand{\slat}{SL\textsc{at}\xspace}
\newcommand{\greencheck}{\textcolor{green!70!black}{\ding{52}}}
\newcommand{\redcross}{\textcolor{red!70!black}{\ding{55}}}

\author{%
  Minkyung Kwon\thanks{: Equal contribution}
  \quad\;
  Jinhyeok Choi\footnotemark[1] 
  \quad
  Youngjin Shin 
  \quad
  Jaeyeong Kim 
  \quad
  JongMin Lee \; \\
  \textbf{Seungryong Kim}\thanks{: Corresponding author} \\[5pt] 
  KAIST AI\\[5pt]
\tt\ \small \textcolor{blue!60!black}{\href{https://cvlab-kaist.github.io/MORPHOS}{https://cvlab-kaist.github.io/MORPHOS}}
}

\begin{document}

\maketitle
\begin{figure}[htbp]
    \centering
    \vspace{-20pt}
    \includegraphics[width=1\textwidth]{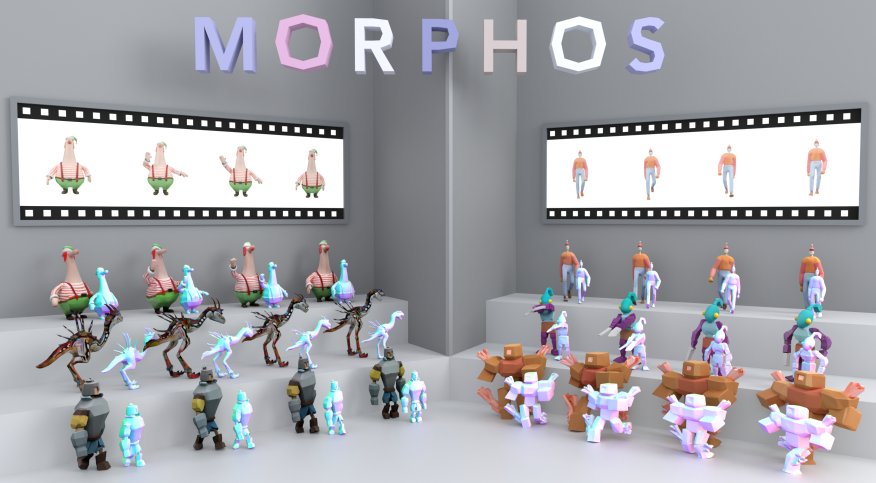}
    \caption{\textbf{Teaser.} Given video inputs, \textbf{\ours} autoregressively generates unified dynamic 3D representations (meshes, 3D Gaussians, and radiance fields).}
    \label{fig:duck}\vspace{-10pt}
\end{figure}

\begin{abstract}

\vspace{-10pt}
We present \ours, a novel autoregressive framework that generates dynamic 3D assets from videos across diverse representations, including meshes, 3D Gaussians, and radiance fields. Existing methods are typically limited to a single representation, struggle to model topological changes, or fail to maintain temporal consistency over long videos. To address these limitations, we introduce the Temporal Structured Latents (\tslat), a unified 4D representation that jointly encodes geometry and appearance along the temporal dimension. Leveraging \tslat, \ours autoregressively generates dynamic 3D assets via causal attention, conditioning each frame on its preceding history to ensure temporal consistency while handling evolving topologies. We also propose a {temporal-structural augmentation} to mitigate error accumulation in autoregressive generation. \ours achieves state-of-the-art performance in appearance and competitive results in geometry across multiple benchmarks, demonstrating superior generalization across various representations and robustness in long-horizon generation.
\end{abstract}

\section{Introduction}


The rapid evolution of 3D generative models has significantly advanced the frontiers of digital content creation, enabling the high-fidelity synthesis of 3D assets from image or text prompts~\cite{li2025triposg,zhao2025hunyuan3d,li2025step1x, xiang2025structured,li2024craftsman3d,zhang2024clay}. These frameworks utilize diverse representations, ranging from meshes for geometry modeling to 3D Gaussians~\cite{kerbl20233d} or Radiance Fields~\cite{mildenhall2021nerf} for volumetric rendering. Building upon these foundations, 4D generative models extend synthesis to the temporal dimension, leveraging video inputs to generate dynamic, animated 3D content~\cite{sabathier2026actionmesh, chen2026motion, ren2024l4gm, zhang2025gaussian, park2021nerfies, yenphraphai2025shapegen4d, yin2026sculpt4d, jiang2026mesh4d}.

\begin{figure}[t]
\centering
\begin{minipage}{0.44\textwidth}
    \centering
    \includegraphics[width=\textwidth]{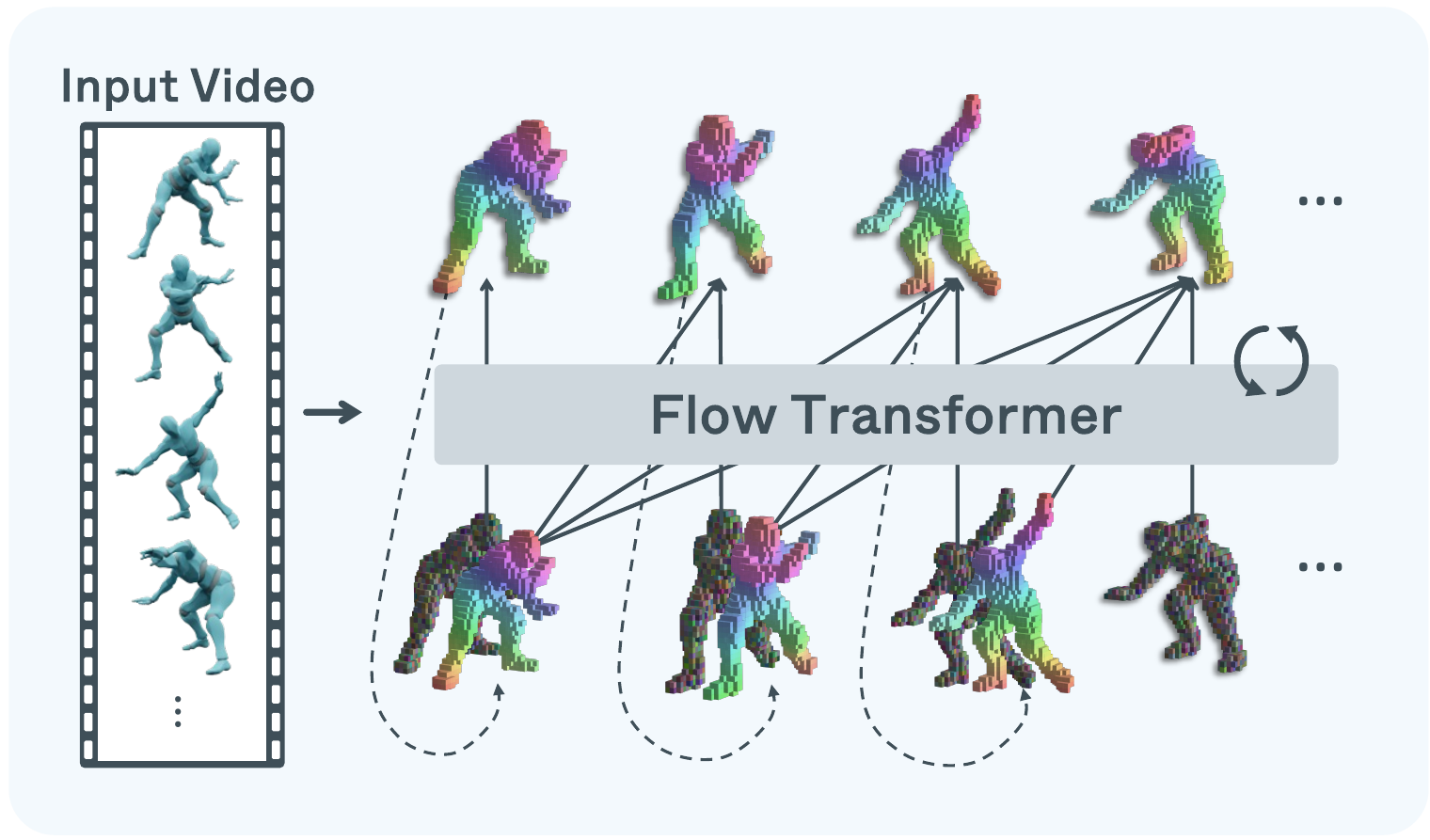}
    \captionof{figure}{\textbf{Autoregressive streaming.}}
    \label{fig:intro_ar}
\end{minipage}
\hfill
\begin{minipage}{0.52\textwidth}
    \centering
    \captionof{table}{\textbf{Baseline comparison.} \ours generates diverse 3D representations, while maintaining temporal consistency in long-horizon generation.}
    \label{tab:intro_comparison}
    
    \fontsize{7}{8.5}\selectfont
    \setlength{\tabcolsep}{4pt}
    \begin{tabular}{l|ccc|c}
        \toprule
        \multirow{2}{*}{\textbf{Method}} & \multicolumn{3}{c|}{\textbf{Representation}} & \textbf{Long-horizon} \\
        & Mesh & Gaussian & RF & \textbf{Generation} \\
        \midrule
        Motion324~\cite{chen2026motion} & \greencheck & \redcross & \redcross & \greencheck \\
        ActionMesh~\cite{sabathier2026actionmesh} & \greencheck & \redcross & \redcross & \redcross \\
        Mesh4D~\cite{jiang2026mesh4d} & \greencheck & \redcross & \redcross & \redcross \\
        L4GM~\cite{ren2024l4gm} & \redcross & \greencheck & \redcross & \greencheck \\
        GVFD~\cite{zhang2025gaussian} & \redcross & \greencheck & \redcross & \redcross \\
        \midrule
        \textbf{MORPHOS (Ours)} & \greencheck & \greencheck & \greencheck & \greencheck \\
        \bottomrule
    \end{tabular}
\end{minipage}
\vspace{-10pt}
\end{figure}

Despite their promise, current 4D generation methods suffer from two primary limitations: \emph{representation fragmentation} and \emph{inability to handle topological changes}. Existing frameworks typically specialize in a single, incompatible 3D format like meshes~\cite{sabathier2026actionmesh,yenphraphai2025shapegen4d,jiang2026mesh4d,chen2026motion} or 3D Gaussians~\cite{ren2024l4gm,zhang2025gaussian}. This specialization restricts model generalization across diverse modalities. Furthermore, they typically rely on deformation-based modeling to maintain temporal consistency~\cite{chen2026motion, sabathier2026actionmesh, jiang2026mesh4d, zhang2025gaussian}. While effective for rigid motions, these methods struggle to accommodate topological changes or significant structural shifts over time.

To address these challenges, we introduce \textbf{\ours}, an autoregressive 4D generative model designed for unified, animated 3D representations. We first propose \tslat (Temporal Structured Latents), a 4D representation which extends 3D structured latent~\cite{xiang2025structured} to the temporal domain. This latent space facilitates the joint encoding of geometry and appearance along the temporal dimension, while capturing complex structural transformations. Our proposed model, \ours, consists of two autoregressive flow models~\cite{liu2022flow, xiang2025structured}: the first generates a sparse structure (\textit{i.e.,} voxel) and the second generates \tslat conditioned on the sparse structure. In both models, we adopt a \textit{causal attention architecture} that conditions each frame on its preceding history, enabling long-horizon 4D generation from videos with large motion and evolving topologies (see~\Cref{fig:intro_ar}). Furthermore, this causal architecture enables key-value (KV) caching~\cite{gao2024ca2, liu2025rolling, yin2025causvid, huang2025selfforcing}, which significantly reduces redundant computations and accelerates inference.


To further ensure temporal consistency during the long-horizon generation, we introduce a \textit{temporal-structural augmentation} strategy during training. Specifically, it addresses the accumulation of errors in the preceding history frames, as well as the imperfect sparse structure used in \tslat generation. The \textit{temporal augmentation} applies independent noise levels across frames during training, exposing the model to histories of varying quality and making it robust to imperfect past context at inference time. In addition, \textit{structural augmentation} perturbs the sparse voxel structures conditioned on the generation of \tslat, ensuring robustness to structural inaccuracies accumulated during the two-stage generation process.
Experimental evaluations across multiple 4D benchmarks~\cite{sabathier2026actionmesh,chen2026motion,jiang2023consistent4d} demonstrate that \ours achieves state-of-the-art performance in appearance and competitive results in geometry. Crucially, the framework effectively models both geometry and appearance even under significant topological transitions during long-horizon generation (see \Cref{tab:intro_comparison}). Our main contributions are as follows:

\begin{enumerate}
    \item We introduce an autoregressive 4D generative framework for unified and dynamic 3D representations, with complex motion and topological changes.
    \item We propose a training strategy with temporal-structural augmentation to enhance the stability and robustness of long-horizon 4D generation.
    \item \ours demonstrates state-of-the-art performance across multiple 4D benchmarks, achieving superior geometric fidelity and appearance with minimal error accumulation.
\end{enumerate}

\section{Related Work}
\subsection{3D Asset Generation}

A growing line of works~\cite{zhang20233dshape2vecset, xiang2025structured,zhang2024clay, li2024craftsman3d, zhao2025hunyuan3d, li2025step1x, li2025triposg, chen2025sam} generate 3D assets from images or text inputs, by utilizing a diffusion process on learned, scalable 3D latent space~\cite{zhang20233dshape2vecset,xiang2025structured}. 3DShape2VecSet~\cite{zhang20233dshape2vecset} introduces unordered, diffusible latent vector sets for encoding meshes and point clouds. TRELLIS~\cite{xiang2025structured} proposes voxel-based structured latents (\slat), which can be decoded into diverse 3D representations — including meshes, 3D Gaussians~\cite{kerbl20233d}, and Radiance Fields~\cite{mildenhall2021nerf}. 
Follow-up works extend this paradigm in various directions~\cite{chen2025ultra3d,he2025sparseflex, xiang2025native,huang2025cupid,huang2026anigen}. Our approach is built upon \slat for robust and unified 4D generation.


\subsection{4D Asset Generation}

Recent 4D generation methods~\cite{ren2024l4gm,zhang2025gaussian,yenphraphai2025shapegen4d,yin2026sculpt4d,sabathier2026actionmesh,chen2026motion,jiang2026mesh4d} extend 3D asset generation to dynamic 3D assets given a video input. Some 3D Gaussian-based methods reconstruct per-frame 3D Gaussians~\cite{ren2024l4gm} or predict the deformation field of 3D Gaussians~\cite{zhang2025gaussian}, but operate entirely in Gaussian space without explicit geometric representations. On the other hand, mesh-based methods~\cite{yenphraphai2025shapegen4d,yin2026sculpt4d,jiang2026mesh4d} typically fine-tune a pretrained 3D generator with temporal attention to produce mesh sequences. Some works utilize a deformation VAE decoder~\cite{sabathier2026actionmesh,jiang2026mesh4d}, or leverage a feed-forward transformer to predict the deformation of the first frame's mesh~\cite{chen2026motion}, producing topology-consistent mesh output. However, deformation-based approaches struggle to model structural transitions. In contrast, we generate frame-wise 3D representations with temporally consistent motion.


\subsection{Autoregressive Video Diffusion}

Early video diffusion models~\cite{wan2025wan, blattmann2023stable, yang2024cogvideox, blattmann2023videoldm, ho2022video} typically employ bidirectional attention to denoise frames simultaneously. Conversely, recent works~\cite{gu2025long,yin2025causvid,chen2025diffusion,huang2025selfforcing,gao2024ca2,jin2024pyramidal,liu2025rolling} adopt autoregressive frameworks for improved scalability and long-horizon generation. Some methods~\cite{gao2024ca2, liu2025rolling, yin2025causvid, huang2025selfforcing} employ sliding temporal windows and causal attention architecture to enable efficient inference via key-value (KV) caching. Diffusion Forcing~\cite{chen2025diffusion, song2025history} utilizes independent per-frame noise levels to mitigate the training-inference gap, where the model is trained on ground truth history but conditioned on self-generated frames with error. While we adopt this formulation, the temporal augmentation alone is insufficient for our framework, as structural errors accumulate during the later stages of the diffusion process, highlighting the necessity of the structural augmentation.
\section{Preliminary: Structured Latents for 3D Generation}
\label{trellis_basics}

Structured latents (\slat)~\cite{xiang2025structured} is a unified latent space that can be decoded into meshes, 3D Gaussians~\cite{kerbl20233d}, and Radiance Fields~\cite{mildenhall2021nerf}. Specifically, a 3D asset $\mathcal O$ is encoded to a \slat $\mathbf z= \{(z_i, x_i)\}^L_{i=1}$, where $x_i \in \{1, ..., N\}^3$ is a coordinate in the 3D voxel grid, and $z_i \in \mathbb R^{C}$ is a visual feature attached on the voxel at $x_i$. Here, $L \ll N^3$ denotes the number of occupied voxels. Specifically, to encode $\mathcal O$, the asset is first voxelized to obtain a sparse structure $\mathbf x = \{x_i\}^L_{i=1}$. Multi-view images are rendered around the asset to extract the visual features from a vision encoder~\cite{oquab2023dinov2}, which are then projected to the voxels $\{x_i\}^L_{i=1}$. 
This gives us a voxelized feature $\mathbf{f} = \{(f_i, x_i)\}_{i=1}^{L}$, where $f_i \in \mathbb{R}^{D}$, is the feature vector at voxel $x_i$. This sparse representation is then encoded by a sparse VAE~\cite{xiang2025structured} into the \slat $\mathbf{z}=\{(z_i, x_i)\}_{i=1}^{L}$, which can be decoded into diverse 3D formats by different decoders. To generate 3D assets, a two-stage rectified flow process~\cite{liu2022flow} is applied to synthesize sparse structure and \slat sequentially.

\paragrapht{Sparse structure generation.}
The model first generates a sparse structure $\mathbf x$ given an input image to capture an object's coarse shape. The sparse structure is converted to a 3D occupancy binary grid $\mathbf o \in \mathbb \{0,1\}^{N \times N \times N}$ and then encoded by VAE~\cite{xiang2025structured} with 3D convolutions to a downsampled feature $\mathbf s \in \mathbb R^{D \times D \times D \times C_s}$. The rectified flow~\cite{liu2022flow} transformer $\mathcal{G}_{\text {S}}$~\cite{xiang2025structured} predicts the velocity field that transports a noise to the feature $\mathbf{s}$ via flow matching. The generated feature $\mathbf{s}$ can be decoded into the occupancy grid $\mathbf{o}$, from which the sparse structure $\mathbf x$ is obtained.


\paragrapht{Structured latent generation.}
Given the active voxels in the generated sparse structure, another rectified flow transformer $\mathcal{G}_{\text{L}}$~\cite{xiang2025structured} generates the \slat $\mathbf{z}=\{(z_i, x_i)\}_{i=1}^{L}$ from a structured noise, for fine-grained geometry and appearance from the input image. Finally, this generated \slat is decoded into meshes, 3D Gaussians, and Radiance Fields with the dedicated decoders~\cite{xiang2025structured}.

\section{Method}

Given an input video $\mathbf{I}^{1:T} = \{\mathbf I^t\}_{t=1}^{T}$, where $\mathbf I^t \in \mathbb{R}^{H \times W \times 3}$ denotes an RGB image at time $t$, our goal is to generate dynamic 3D assets including meshes, 3D Gaussians~\cite{kerbl20233d}, and Radiance Fields~\cite{mildenhall2021nerf}. For a unified modeling of both geometry and appearance, we leverage a pre-trained image-to-3D generative model with structured 3D latents~\cite{xiang2025structured}. Building upon this, we design three key components: (i) \tslat, which jointly encodes geometry and appearance along the temporal dimension (\Cref{sec:method_tslat}); (ii) autoregressive generation with causal architecture, enabling efficient inference (\Cref{sec:method_ar}); (iii) temporal-structural augmentation training strategy to mitigate error accumulation for robust long-horizon generation (\Cref{sec:method_aug}).

\subsection{Temporal Structured Latents for 4D Generation}
\label{sec:method_tslat}
\begin{wrapfigure}{r}{0.4\textwidth}
    \vspace{-20pt}
    \centering
    \includegraphics[width=0.4\textwidth]{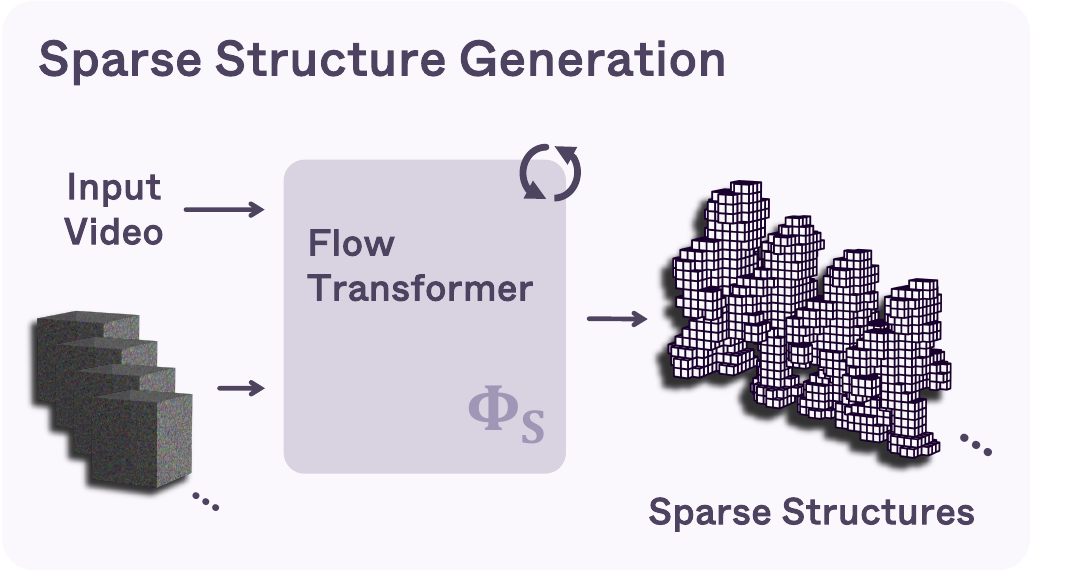}
    \caption{\textbf{Sparse structure generation.}}
    \label{fig:ss_architecture}
    \vspace{-10pt}
\end{wrapfigure} 

We introduce \tslat (Temporal Structured Latents) by extending \slat into the temporal domain, which can be decoded into versatile dynamic 3D representations. We define an animated 3D asset, $\mathcal O^{1:T} = \{\mathcal O^t\}_{t=1}^T$, where $\mathcal O^t=\langle \mathcal V^t, \mathcal F^t, \mathcal T^t\rangle$ is a mesh at time $t$ with vertices $\mathcal V^t$, faces $\mathcal F^t$, and texture $\mathcal T^t$.
We then encode this mesh sequence into the \tslat $\mathbf{z}^{1:T} = \{\mathbf{z}^t\}_{t=1}^{T}$, where $\mathbf{z}^t=\{(z_i^t, x_i^t)\}_{i=1}^{L_t}$ is \slat at time $t$. To preserve the temporal coherence of the animated 3D asset, we employ a global spatial normalization before the \tslat encoding. We compute a global union Axis-Aligned Bounding Box (AABB) across the entire sequence of length $T$.
Specifically, given the animated 3D asset $\mathcal O^{1:T}$, we determine the global spatial extrema across all frames. We define the sequence-level bounding box by extracting the minimum and maximum coordinates along each spatial axis $j \in \{x, y, z\}$:
\begin{equation}
v_{\min}^{(j)} = \min_{1 \le t \le T} \Big( \min_{v \in \mathcal V^t} v^{(j)} \Big), \quad v_{\max}^{(j)} = \max_{1 \le t \le T} \Big( \max_{v \in \mathcal V^t} v^{(j)} \Big)
\end{equation}
where $v_{\min}, v_{\max} \in \mathbb R^3$. From these bounds, we derive a single scene center $c = \frac{1}{2}(v_{\min} + v_{\max})$ and a scale factor $s = 1 / \max_j(v_{\max}^{(j)} - v_{\min}^{(j)})$. For every frame $t$, the mesh vertices are transformed via $\bar{\mathcal V}^t = s(\mathcal V^t - c)$, confining the entire motion sequence within a normalized canonical space.
\subsection{Autoregressive 4D generation}
\label{sec:method_ar}
\begin{figure}[t]
    \centering
    \includegraphics[width=\textwidth]{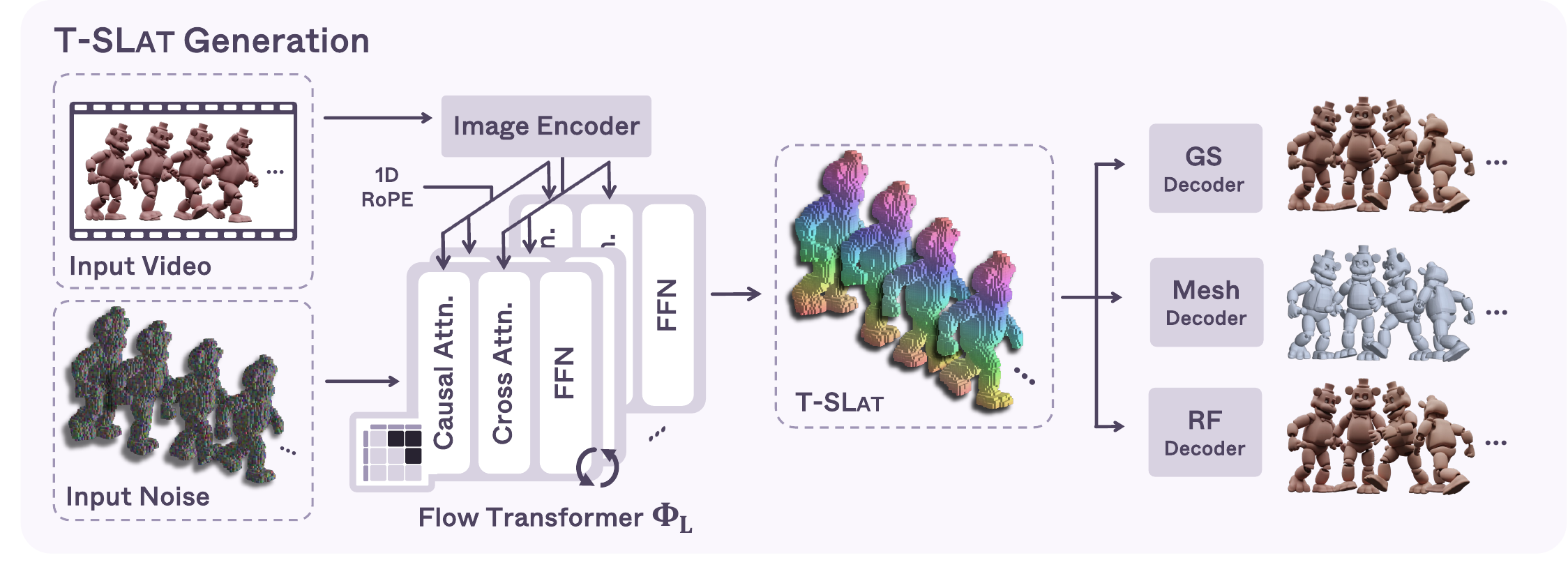}
    \vspace{-10pt}
    \caption{\textbf{\tslat generation.} \tslat provides a unified representation for dynamic 3D generation. To enable AR generation, we employ a causal attention architecture over frames. The temporal augmentation applies different noise timesteps across frames for robust long-horizon modeling, while spatial augmentation uses voxel dropout to improve robustness against structural errors.}
    \label{fig:slat_architecture}
    \vspace{-15pt}
\end{figure}
    
    


To synthesize dynamic 3D assets from input videos, we propose a two-stage autoregressive generative framework within the \tslat space. This approach decomposes 4D generation into the sequential synthesis of sparse structure features $\mathbf{s}^{1:T}=\{\mathbf{s}^t\}_{t=1}^T$ and temporal structured latents $\mathbf{z}^{1:T}$. Specifically, we factorize the joint distribution of the \tslat as a Markovian process:
\begin{equation}
p(\mathbf{z}^{1:T} \mid \mathbf{I}^{1:T}) = \prod_{t=1}^{T} p(\mathbf{z}^t \mid \mathbf{z}^{<t}, \mathbf I^t),
\end{equation}
where each latent $\mathbf{z}^t$ is conditioned on the preceding frames and the corresponding input video frame $\mathbf I^t$. This factorization facilitates 4D generation across arbitrary horizons by modeling the latent distribution sequentially, effectively removing the requirement for full sequence access during generation. In practice, we employ a local window to restrict the maximum length of history frames.

\paragrapht{Flow transformers with causal attention.}
To model the conditional distributions defined in our Markovian factorization, we utilize two rectified flow~\cite{liu2022flow} transformers, $\Phi_{\text S}$ and $\Phi_{\text L}$. These models predict a velocity field from a noisy sample to the target distribution through a flow matching~\cite{lipman2022flow}. Video conditions are encoded via a pre-trained visual encoder~\cite{oquab2023dinov2} and conditioned into the model through cross-attention layers.
\begin{wrapfigure}{r}{0.3\textwidth}
    \vspace{-10pt}
    \centering
    \includegraphics[width=\linewidth]{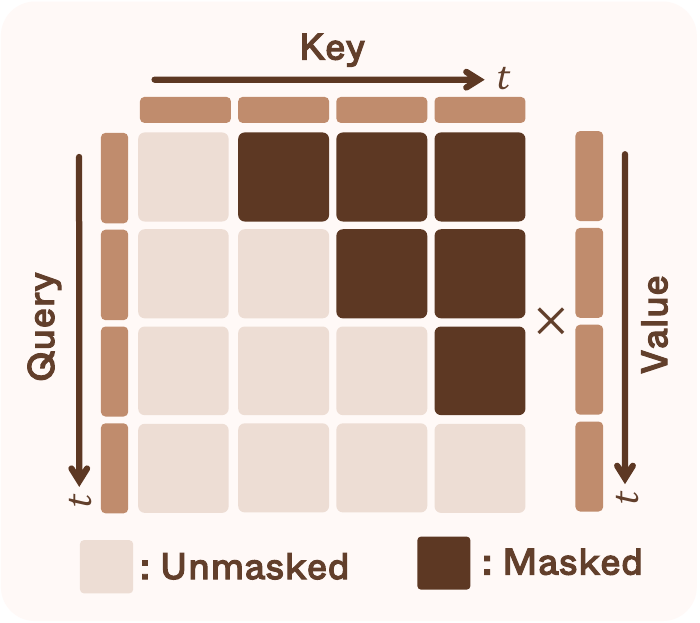}
    \caption{\textbf{Causal attention.}}
    \label{fig:causal_mask}
\end{wrapfigure}
We implement an autoregressive generation process using a causal attention architecture with the local window of size $w$. This architecture ensures that the receptive field for a query at frame $t$ is strictly limited to the preceding history frames within the local window.



To capture the temporal ordering of token sequences, we apply 1D Rotary Position Embeddings (RoPE)~\cite{su2024roformer} along the temporal dimension. During inference, we utilize key-value (KV) caching to store the key and value tokens of previously generated frames. The KV cache is dynamically managed to retain a total of $w$ frames, ensuring the sliding window design. We cache them before the RoPE, which is computed on-the-fly to ensure accurate relative distance as the window slides through the sequence.

\subsection{Training with Temporal-Structural Augmentation}
\label{sec:method_aug}
We optimize our two flow transformers with rectified flow training~\cite{liu2022flow}, to learn to transport a source noise distribution $p_1 = \mathcal{N}(0, I)$ to the target data distribution $p_0$. Then the intermediate latent $\mathbf{y}_k$ at noise level $k \in [0, 1]$ is constructed via linear interpolation:
\begin{equation}
\mathbf{y}_k = (1 - k)\mathbf{y}_0 + k\epsilon, \quad \epsilon \sim \mathcal{N}(0, I),
\end{equation}
where $\mathbf{y}_0$ denotes the ground-truth latent. Accordingly, the transformers $\Phi_{\text S}$ and $\Phi_{\text L}$ are trained to predict the corresponding velocity field $v_{\Phi_\text S}(\mathbf{s}^t_k, k)$ and $v_{\Phi_\text L}(\mathbf{z}^t_k, k)$,  which points from the noise toward the clean data. In our conditional distributions, this velocity field is conditioned on the previous history and the current video frame. However, autoregressive models often suffer from a training inference discrepancy~\cite{huang2025selfforcing} because they are trained on ground-truth history but must generalize to self-generated, imperfect history during inference. To bridge this gap, we introduce \textit{temporal-structural augmentation} strategy during training.

\paragrapht{Temporal augmentation.}
To mitigate error accumulation from self-generated history, we implement temporal augmentation inspired by Diffusion Forcing~\cite{chen2025diffusion}. During training, rather than applying a single noise level to the entire frame sequence, we assign an independent noise level from a uniform distribution $k_t \sim \mathcal{U}(0, 1)$ to each frame $t$ within the temporal window. This approach compels the model to learn robust sequential dependencies by denoising the current frame $t$ while conditioned on preceding noisy frames. Consequently, the model becomes resilient to the cumulative errors in autoregressive rollouts.

\paragrapht{Structural augmentation.}
While temporal augmentation stabilizes the sequence over time, the latent stage $\Phi_{\text{L}}$ remains sensitive to inaccuracies in the sparse structure $\mathbf{s}$ generated by the preceding stage $\Phi_{\text{S}}$. To align the training distribution with inference, we employ structural augmentation by perturbing voxel structure. For given $\mathbf{z}^t_{k_t}$, we randomly drop voxels with a probability $\lambda$ to obtain $\tilde{\mathbf{z}}^t = \{ (z_i, x_i) \}_{i \in \mathcal{S}}$ during the training of $\Phi_{\text{L}}$, where  $\mathcal{S} \subseteq \{1, \dots, L\}$ is a subset of voxel indices. Same augmentation is applied to ground-truth history $\mathbf z^{<t}$ and target latent $\mathbf z^t_0$. This perturbation encourages the model to synthesize high fidelity \tslat features from incomplete sparse structures, effectively alleviating structural errors accumulated through the structure generation stage.

\paragrapht{Training objective.}
We train both transformers of \ours on rectified flow matching loss~\cite{liu2022flow}. The loss for the sparse structure generator $\Phi_{\text{S}}$ is defined as:
\begin{equation}
\mathcal{L}_{\text{S}} = \mathbb{E}_{t, k_t, \epsilon} \left[ \| v_{\Phi_{\text{S}}}(\mathbf{s}^t_{k_t}, k_t, \mathbf I^t, \mathbf{s}^{<t}) - (\mathbf{s}^t_0 - \epsilon) \|_2^2 \right].
\end{equation}
Similarly, the \tslat generator $\Phi{_\text{L}}$ is optimized with structural augmentation using perturbed latents $\tilde{\mathbf{z}}^t_{k_t}$, $\mathbf z^{<t}$, and $\tilde{\mathbf{z}}^t_0$:
\begin{equation}
\mathcal{L}_{\text{L}} = \mathbb{E}_{t, k_t, \epsilon} \left[ \| v_{\Phi_{\text{L}}}(\tilde{\mathbf{z}}^t_{k_t}, k_t, \mathbf I^t, \tilde{\mathbf{z}}^{<t}) - (\tilde{\mathbf{z}}^t_0 - \epsilon) \|_2^2 \right].
\end{equation}


\section{Experiments}
\label{sec:experiments}
\subsection{Implementation Details}  
\paragraph{Training dataset.}
We construct 10K animated 3D assets and render 12-frame videos from Objaverse~\cite{objaverse} and Objaverse-XL~\cite{objaverseXL}. We compute a union axis-aligned bounding box (AABB) across all frames of a scene, to globally normalize the entire scene into a canonical $[-0.5, 0.5]^3$ coordinate space. For the video inputs, we render RGBA images with $512\times512$ resolution. To augment the dataset, we render multiple video inputs for each 3D asset with varying camera positions and FoVs. To ensure motion qualities, we apply a filtering algorithm based on the sub-mesh extent ratio and the number of occupied voxels. Additional details of the training dataset are provided in~\Cref{appendix:dataset}.
 
\paragraph{Training details.}
We build \ours upon a pre-trained 3D generative model~\cite{xiang2025structured}. Specifically, we inflate the self-attention layers to 3D self-attention by concatenating each frame tokens into a single sequence, and apply causal mask. To maintain robust 3D generative prior of the pre-trained model, we fine-tune only causal attention and AdaLN parameters. During training, we use a local window of size $w=3$ and structural augmentation with $\lambda=0.05$. Both models are optimized with AdamW~\cite{loshchilov2017decoupled} using a fixed learning rate of $1 \times 10^{-4}$ and an EMA decay of $0.9999$. We train $\Phi_{\text{S}}$ and $\Phi_{\text{L}}$ for 200k iterations with batch sizes of $32$ and $64$, respectively, on $2$ NVIDIA B200 GPUs. Additional training details are provided in~\Cref{appendix:training_details}.
 \begin{figure}[t]
    \centering
    \includegraphics[width=\textwidth]{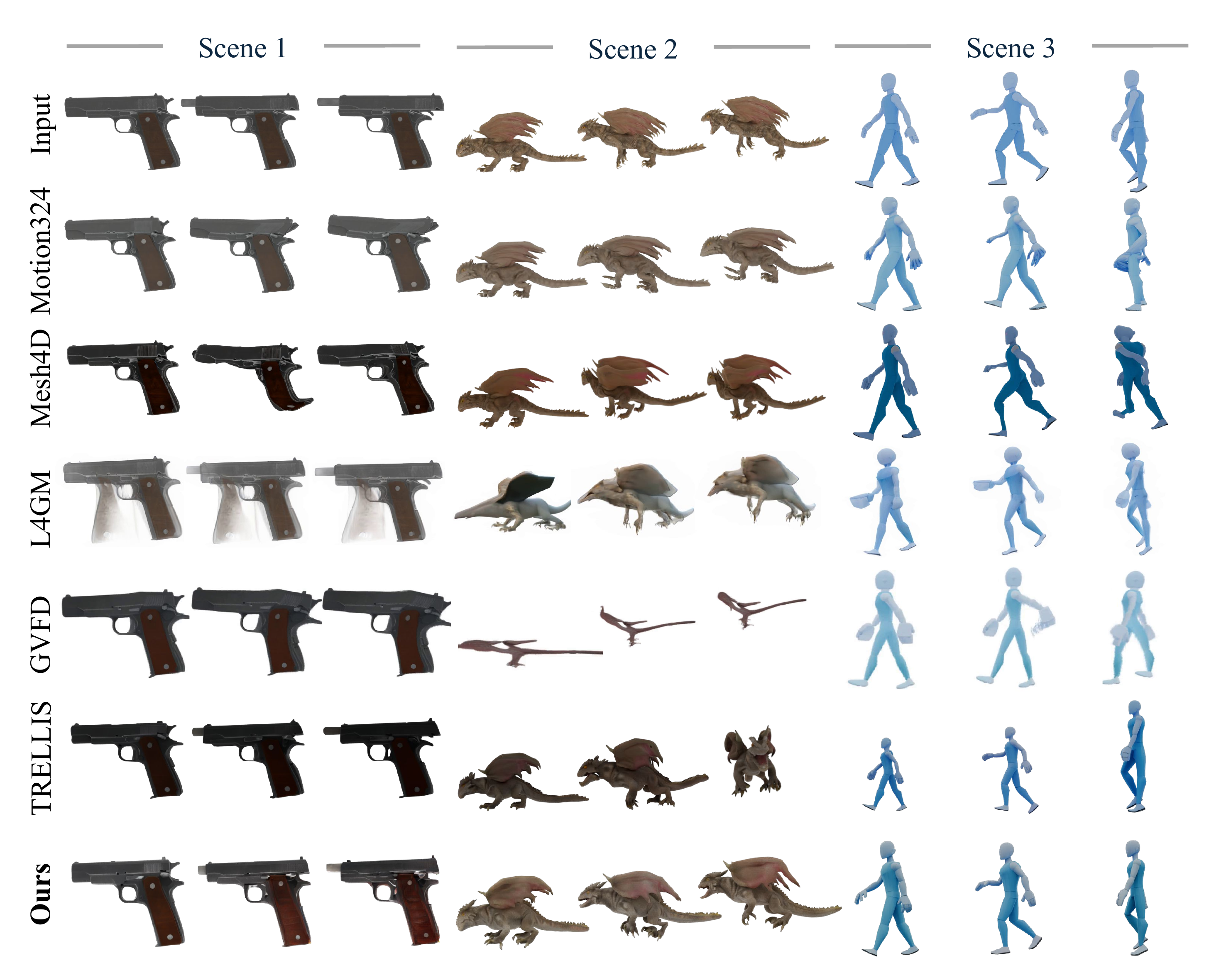}
    \caption{\textbf{Qualitative results on appearance.} \ours  produces visually consistent and high-fidelity appearance, maintaining stable textures throughout the entire sequence.}
    \label{fig:qual_appearance}
    \vspace{-10pt}
\end{figure}
\subsection{Evaluation} 


We evaluate \ours on both geometry and appearance, against state-of-the-art video-to-4D generation baselines~\cite{chen2026motion,sabathier2026actionmesh,jiang2026mesh4d,zhang2025gaussian,ren2024l4gm} as well as the image-to-3D model TRELLIS~\cite{xiang2025structured}. We use mesh for geometry and 3D Gaussians for appearance evaluation. Detailed evaluation protocols are provided in~\Cref{appendix:inference_details}, and the details of evaluation metrics are included in~\Cref{appendix:appearance_eval,appendix:geometric_eval}.

\begin{figure}[t]
    \centering
    \includegraphics[width=\textwidth]{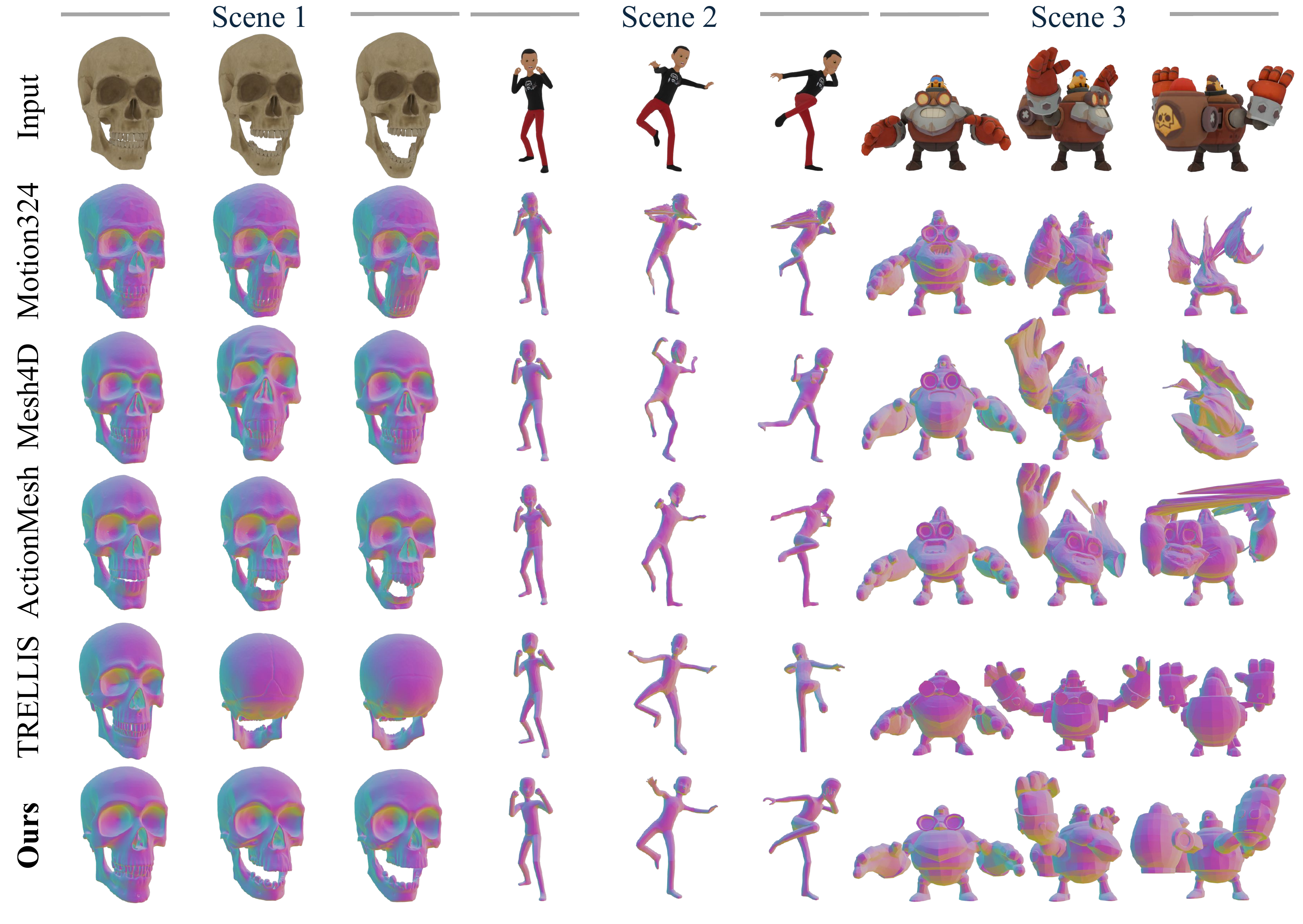}
    \vspace{-10pt}
    \caption{\textbf{Qualitative results on geometry.} \ours produces stable and coherent geometry, maintaining accurate shape and temporal consistency throughout the sequence.}
    \label{fig:qual_geo}
    \vspace{-11pt}
\end{figure}

\paragrapht{Evaluation benchmark.}
We evaluate our method on three benchmarks: (i) Motion80~\cite{chen2026motion}, which includes 64 short sequences and 16 long sequences, where long sequences exceed 128 frames; (ii) ActionBench~\cite{sabathier2026actionmesh}, which consists of 128 animated scenes with 16 frames each; and (iii) Consist4D~\cite{jiang2023consistent4d}, which contains 7 videos, each with 32 frames. Since both Motion80 and ActionBench provide ground-truth meshes, we evaluate both geometry and appearance quality. Consist4D only provides input and ground-truth videos, thus we evaluate only appearance quality following azimuth alignment.
For appearance evaluation, we use LPIPS~\cite{zhang2018unreasonable}, CLIP similarity~\cite{radford2021learning}, DreamSim~\cite{fu2023dreamsim}, and Fréchet Video Distance (FVD)~\cite{unterthiner2019fvd}. For geometry evaluation of meshes, we measure Chamfer Distance (CD), F-score, and Point-to-Surface distance (P2S). Since ActionBench provides no ground-truth mesh surface, we measure CD and F-score only on ActionBench.


\paragrapht{Baselines.}
To evaluate 4D generation performance, we compare our method with: (i) video-to-4D mesh generation models, including Motion324~\cite{chen2026motion}, ActionMesh~\cite{sabathier2026actionmesh}, and Mesh4D~\cite{jiang2026mesh4d}; (ii) image-to-3D model, TRELLIS~\cite{xiang2025structured}, which we run independently on each frame; and (iii) video-to-4D Gaussian generation models, including L4GM~\cite{ren2024l4gm} and GVFD~\cite{zhang2025gaussian}.
ActionMesh~\cite{sabathier2026actionmesh} is excluded from appearance evaluation as it does not produce texture outputs. For \ours evaluation, we use KV caching by default for both $\Phi_{\text{S}}$ and $\Phi_{\text{L}}$.

\subsection{Results}

\begin{figure}[t]
\centering
\fontsize{6.5}{9}\selectfont

\begin{minipage}[t]{0.58\textwidth}
\centering
\setlength{\tabcolsep}{2.0pt}

\captionof{table}{\textbf{Quantitative evaluation on Motion80~\cite{chen2026motion}.}}
\label{tab:mesh_motion80}

\begin{tabular}{c|ccc|c|ccc}
\toprule

& \multicolumn{3}{c}{\textbf{Appearance}} 
& \multicolumn{1}{c}{\textbf{Video}}
& \multicolumn{3}{c}{\textbf{Geometry}} \\
\cmidrule(lr){2-4} \cmidrule(lr){5-5} \cmidrule(lr){6-8}

\textbf{Method} & LPIPS$\downarrow$ & CLIP$\uparrow$ & DreamSim$\downarrow$
& FVD$\downarrow$
& CD$\downarrow$& F-score$\uparrow$& P2S$\downarrow$ \\
\midrule

\multicolumn{8}{c}{\textbf{Short}} \\
\midrule
Motion324~\cite{chen2026motion} & 0.2118 & 0.8051 & 0.2347 & 336.63 & \textbf{0.0615} & \textbf{0.3259} & \textbf{0.0308} \\
ActionMesh~\cite{sabathier2026actionmesh} & - & - & - & - & 0.1062 & \underline{0.2597} & 0.0528 \\
Mesh4D~\cite{jiang2026mesh4d} & 0.2023 & 0.7186 & 0.3465 & 592.56 & 0.1791 & 0.0712 & 0.0813 \\
TRELLIS~\cite{xiang2025structured} & 0.2031 & 0.8643 & 0.1861 & 796.51 & 0.2033 & 0.1354 & 0.1022 \\
L4GM~\cite{ren2024l4gm} & \textbf{0.1296} & \underline{0.8663} & \underline{0.1605} & \textbf{188.32} & - & - & - \\
GVFD~\cite{zhang2025gaussian} & 0.1661 & 0.8439 & 0.1998 & 328.14 & - & - & - \\
\textbf{\ours\ (Ours)} & \underline{0.1505} & \textbf{0.8751} & \textbf{0.1512} & \underline{246.22} & \underline{0.0761} & 0.1455 & \underline{0.0320} \\

\midrule
\multicolumn{8}{c}{\textbf{Long}} \\
\midrule
Motion324~\cite{chen2026motion} & 0.2347 & 0.7905 & 0.2407 & 889.93 & \textbf{0.0701} & \textbf{0.2335} & \underline{0.0353} \\
ActionMesh~\cite{sabathier2026actionmesh} & - & - & - & - & 0.1614 & \underline{0.1718} & 0.0786 \\
Mesh4D~\cite{jiang2026mesh4d} & 0.2408 & 0.5860 & 0.5170 & 1327.54 & 0.4724 & 0.0265 & 0.2250 \\
TRELLIS~\cite{xiang2025structured} & 0.2118 & 0.8359 & 0.2005 & 1527.19 & 0.2383 & 0.0875 & 0.1177 \\   
L4GM~\cite{ren2024l4gm} & \textbf{0.1355} & \underline{0.8578} & \underline{0.1535} & 487.44 & - & - & - \\
GVFD~\cite{zhang2025gaussian} & 0.1796 & 0.8049 & 0.2319 & 827.03 & - & - & - \\
\textbf{\ours\ (Ours)} & \underline{0.1494} & \textbf{0.8670} & \textbf{0.1526} & \textbf{330.59} & \underline{0.0792} & 0.1371 & \textbf{0.0350} \\

\bottomrule
\end{tabular}

\end{minipage}
\hfill
\begin{minipage}[t]{0.38\textwidth}

\centering
\vspace{0pt}
\includegraphics[width=\linewidth]{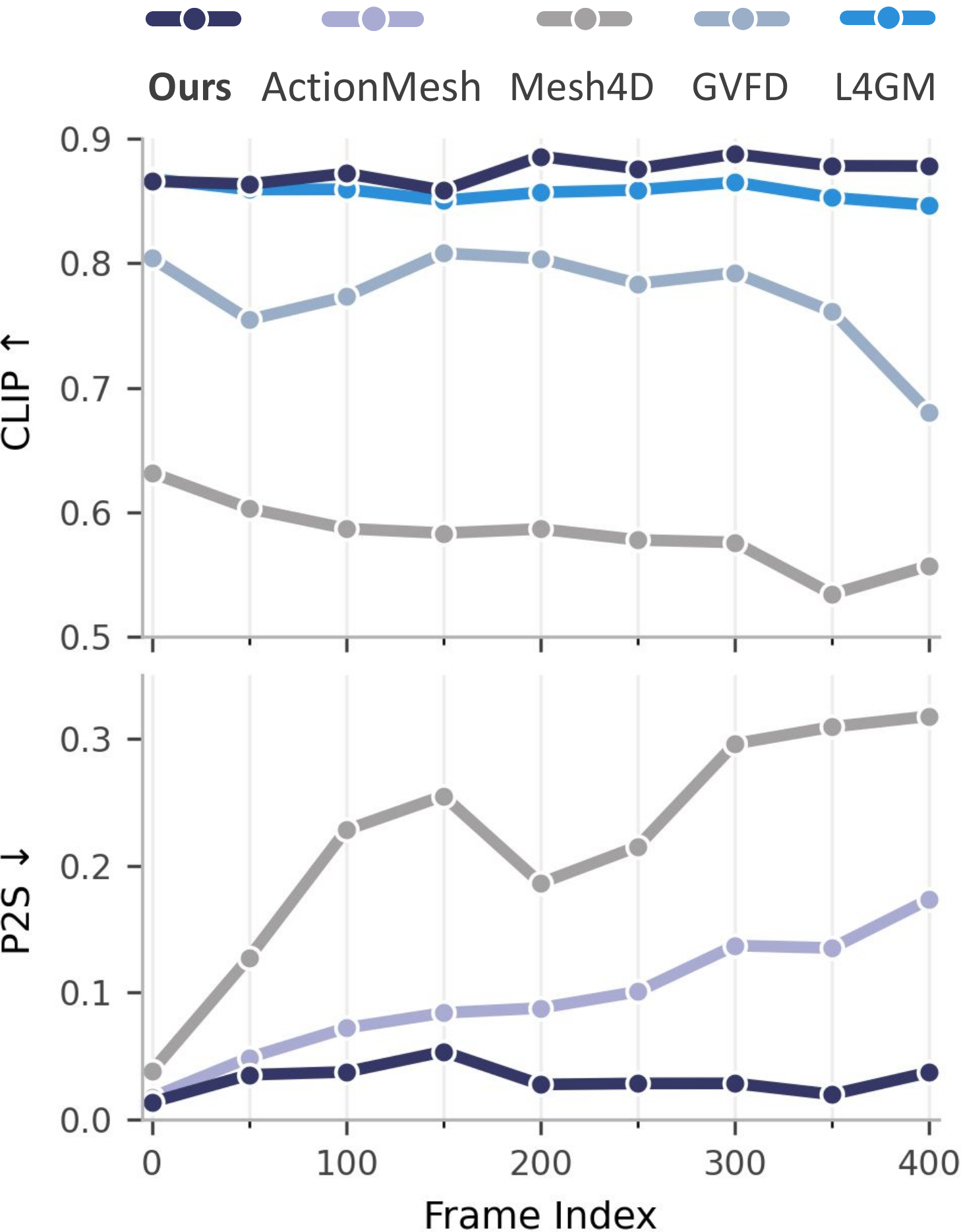}

\captionof{figure}{\textbf{Error accum. analysis.}}
\label{fig:error_accum}

\end{minipage}
\vspace{-2pt}
\end{figure}

\paragrapht{Quantitative results.} We report quantitative results in~\Cref{tab:consist4d_all,tab:mesh_actionbench,tab:mesh_motion80}. For per-frame appearance, \ours achieves state-of-the-art performance across almost all benchmarks~\cite{chen2026motion,jiang2023consistent4d,sabathier2026actionmesh}. In particular, \ours obtains the highest CLIP score across all, demonstrating superior semantic alignment and visual fidelity. For perceptual similarity metrics such as DreamSim and LPIPS, \ours consistently ranks among the top-performing methods, achieving either the best or second-best performance. 

Regarding video appearance, \ours achieves the state-of-the-art FVD score on Motion80-long~\cite{chen2026motion}, as detailed in~\Cref{tab:mesh_motion80}. Notably, Motion80-long contains significantly longer and more complex temporal dynamics than other benchmarks, making long-horizon consistency particularly challenging to maintain. The superior performance of \ours demonstrates the robustness of our autoregressive generation and temporal-structural augmentation when handling extended motions and evolving topologies.
 

In geometry evaluations, \ours achieves competitive results, ranking second on both the Motion80 and ActionBench benchmarks. Although \ours is a unified framework modeling both geometry and appearance, it maintains geometric fidelity comparable to baselines specialized for mesh generation~\cite{sabathier2026actionmesh,chen2026motion}. Standard geometry metrics focus on mesh reconstruction quality, which inherently privileges methods designed solely for mesh generation. Nevertheless, the performance of \ours highlights its robustness and generalization across diverse dynamic 3D representations.

\begin{table*}[t]
\centering
\fontsize{6.5}{7.5}\selectfont

\begin{minipage}[t]{0.58\textwidth}

\centering

\setlength{\tabcolsep}{3pt}

\caption{\textbf{Quantitative evaluation on ActionBench~\cite{sabathier2026actionmesh}.}}

\label{tab:mesh_actionbench}

\begin{tabular}{c|ccc|c|cc}

\toprule

& \multicolumn{3}{c}{\textbf{Appearance}} 

& \multicolumn{1}{c}{\textbf{Video}}

& \multicolumn{2}{c}{\textbf{Geometry}} \\

\cmidrule(lr){2-4} \cmidrule(lr){5-5} \cmidrule(lr){6-7}

\textbf{Method} & LPIPS$\downarrow$ & CLIP$\uparrow$ & DreamSim$\downarrow$

& FVD$\downarrow$

& CD$\downarrow$ & F-score$\uparrow$ \\

\midrule

Motion324~\cite{chen2026motion} & 0.2025 & 0.8304 & 0.2257 & \underline{195.25} & 0.1082 & 0.2013 \\

ActionMesh~\cite{sabathier2026actionmesh} & - & - & - & - & \textbf{0.0898} & \textbf{0.2146} \\


Mesh4D~\cite{jiang2026mesh4d} & \underline{0.1700} & 0.8114 & 0.2447 & 403.20 & 0.1776 & 0.1235 \\

TRELLIS~\cite{xiang2025structured} & 0.2005 & \underline{0.8367} & \underline{0.2199} & 547.21 & 0.1903 & 0.1367 \\

L4GM~\cite{ren2024l4gm} & 0.1908 & 0.8071 & 0.2522 & 211.55 & - & - \\

GVFD~\cite{zhang2025gaussian} & \textbf{0.1687} & 0.8301 & 0.2335 & \textbf{188.67} & - & - \\

\textbf{\ours (Ours)} & 0.1904 & \textbf{0.8551} & \textbf{0.1857} & 
203.02 & \underline{0.0972} & \underline{0.2138} \\

\bottomrule

\end{tabular}

\end{minipage}
\hfill
\begin{minipage}[t]{0.4\textwidth}
\centering
\setlength{\tabcolsep}{2.0pt}
\captionof{table}{\textbf{Evaluation on Consist4D~\cite{jiang2023consistent4d}.}}
\label{tab:consist4d_all}
\begin{tabular}{c|ccc|c}
\toprule
& \multicolumn{3}{c}{\textbf{Appearance}} & \textbf{Video} \\
\cmidrule(lr){2-4} \cmidrule(lr){5-5}
\textbf{Method} & LPIPS$\downarrow$ & CLIP$\uparrow$ & DreamSim$\downarrow$ & FVD$\downarrow$ \\
\midrule
Motion324~\cite{chen2026motion}
& 0.2044 & 0.8285 & 0.2013 & 936.68 \\
Mesh4D~\cite{jiang2026mesh4d}
& 0.1769 & 0.7968 & 0.2507 & 1189.67 \\
TRELLIS~\cite{xiang2025structured}
& 0.2479 & 0.8044 & 0.2962 & 1488.32 \\
L4GM~\cite{ren2024l4gm}
& 0.1633 & \underline{0.8374} & 0.2063 & \underline{825.64} \\  
GVFD~\cite{zhang2025gaussian}
& \textbf{0.1487} & 0.8142 & \textbf{0.1706} & \textbf{821.75} \\
\midrule
\textbf{\ours (Ours)}
& \underline{0.1531} & \textbf{0.8571} & \underline{0.1849} & 1013.11 \\
\bottomrule
\end{tabular}
\end{minipage}
\vspace{-16pt}
\end{table*}
\paragrapht{Qualitative results.} 
\Cref{fig:qual_appearance} presents a qualitative comparison of appearance between \ours and the baselines~\cite{chen2026motion,jiang2026mesh4d,zhang2025gaussian,ren2024l4gm, xiang2025structured}. Deformation-based approaches~\cite{chen2026motion,jiang2026mesh4d,zhang2025gaussian} are inherently restricted by fixed-topology constraints, which prevent them from modeling complex topological transitions. While the reconstruction-based L4GM~\cite{ren2024l4gm} accommodates varying topologies, it suffers from pronounced ghosting artifacts and blurred textures, as observed in Scene 1. Furthermore, TRELLIS~\cite{xiang2025structured} fails to preserve temporal consistency due to its independent frame-wise generation process. In contrast, \ours produces high-fidelity, temporally coherent appearances while seamlessly handling evolving topologies.

\Cref{fig:qual_geo} presents the rendered normal maps of the generated mesh sequences, highlighting the capability of \ours\ to handle complex topological transitions. Deformation-based methods~\cite{chen2026motion,sabathier2026actionmesh,jiang2026mesh4d} struggle to reconstruct sequences involving substantial topological modifications due to their fixed-topology constraints. For instance, they fail to disentangle the upper and lower jaw structures in scene 1, and they introduce severe geometric distortion during the twisting motion in scene 3. Meanwhile, TRELLIS~\cite{xiang2025structured} fails to maintain temporal consistency because of its independent frame-by-frame generation process, which operates without a shared canonical space. In contrast, \ours\ generates geometrically faithful structures, successfully preserving temporal coherence while accommodating intricate structural transitions.

\subsection{Ablation studies}
\label{sec:ablations}

\begin{wraptable}{r}{0.60\textwidth}
\vspace{-10pt}
\centering
\fontsize{7}{7}\selectfont
\setlength{\tabcolsep}{2pt}
\caption{\textbf{Ablation studies on ActionBench~\cite{sabathier2026actionmesh}.}}
\label{tab:ablation_main}
\begin{tabular}{ll|ccc|c|cc}
\toprule
& & \multicolumn{3}{c}{\textbf{Appearance}} & \multicolumn{1}{c}{\textbf{Video}} & \multicolumn{2}{c}{\textbf{Geometry}} \\
\cmidrule(lr){3-5} \cmidrule(lr){6-6} \cmidrule(lr){7-8}
& \textbf{Component} 
& LPIPS $\downarrow$ & CLIP $\uparrow$ & DreamSim $\downarrow$ 
& FVD $\downarrow$ 
& CD $\downarrow$ & F-score $\uparrow$ \\
\midrule
(a) & w/o Causal attn.
& 0.1578 & \underline{0.8487} & 0.1979 
& \underline{323.20} 
& 0.1305 & 0.1986 \\
(b) & w/o Temporal aug.
& 0.1668 & 0.8415 & 0.2084 
& 424.43 
& 0.1291 & 0.1909 \\
(c) & w/o Structural aug.
& 0.1670 & 0.8399 & 0.2087 
& 397.78
& {0.1294} & {0.1906} \\
\multirow{2}{*}{(d)} & w/o $\Phi_{\text{L}}$ training
& \textbf{0.1569} & \textbf{0.8503} & \textbf{0.1956} 
& 370.87 
& \textbf{0.1209} & \underline{0.2026} \\
& w/o $\Phi_{\text{S}}$ training
& 0.1670 & 0.8400 & 0.2104 
& 506.20 
& 0.1770 & 0.1553 \\
\midrule
& \textbf{\ours\ (Ours)}
& \underline{0.1576} & {0.8450} & \underline{0.1966} 
& \textbf{321.37} 
& \underline{0.1219} & \textbf{0.2088} \\
\bottomrule
\end{tabular}
\vspace{-10pt}
\end{wraptable}

We investigate the impact of four core components through ablation studies: (a) the causal attention architecture, (b) temporal augmentation, (c) structural augmentation, and (d) fine-tuning only a single flow model instead of optimizing both $\Phi_{\text{S}}$ and $\Phi_{\text{L}}$. All variants are trained for 10k iterations and evaluated on ActionBench~\cite{sabathier2026actionmesh}. Key-value (KV) caching is enabled by default across all configurations, except (a) bidirectional architecture.

\paragrapht{Causal attention.}
One can adopt a bidirectional attention architecture to perform autoregressive generation via chunk-wise generation with overlapping frames. However, as demonstrated in \Cref{tab:ablation_main}(a), ours causal attention mechanism outperforms the bidirectional baseline across both geometry and appearance metrics. Furthermore, it retains the computational efficiency afforded by key-value (KV) caching. These results demonstrate that the causal attention architecture effectively models the joint distribution required for robust autoregressive generation.

\paragrapht{Temporal augmentation.}
\Cref{tab:ablation_main} (b) presents the ablation analysis of our temporal augmentation strategy. Excluding this strategy leads to degraded performance, as the model can neither effectively model clean-to-noisy interactions during history-conditioned generation, nor maintain robustness against errors within the self-generated history. In contrast, temporal augmentation successfully addresses both challenges, enabling stable next-frame generation while mitigating the error accumulation in autoregressive sampling.

\paragrapht{Structural augmentation.}
\Cref{tab:ablation_main} (c) evaluates the structural augmentation strategy by setting $\lambda=0.0$. Excluding this augmentation marginally improves geometry performance but degrades both per-frame and video appearance metrics. This trade-off indicates that structural augmentation successfully mitigates  propagation from sparse structure generation $\Phi_{\text{S}}$ to \tslat generation $\Phi_{\text{L}}$. Since $\Phi_{\text{L}}$ generates the visual features responsible for appearance, applying our structural augmentation during $\Phi_{\text{L}}$ training effectively stabilizes appearance synthesis.

\begin{figure}[t]
    \centering
    \includegraphics[width=\textwidth]{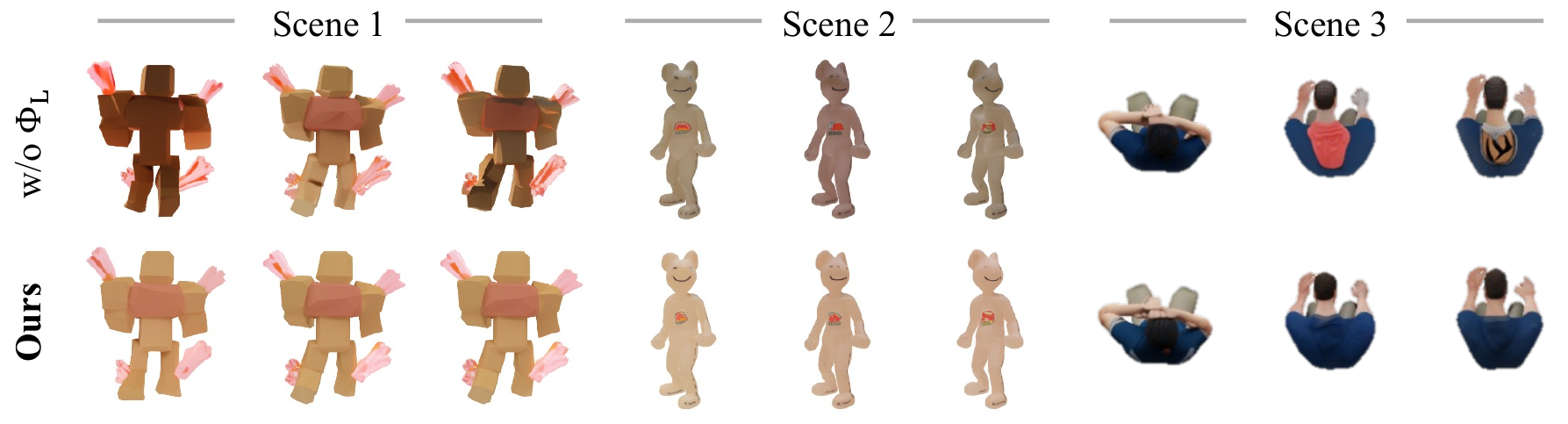}
    \vspace{-10pt}
    \caption{\textbf{Ablation on `w/o $\Phi_\text{L}$ training'.} Without $\Phi_\text{L}$ training, \ours leads to noticeable color degradation and texture inconsistency.}
    \label{fig:appendix_ss_only}
    \vspace{-10pt}
\end{figure}

\paragrapht{Stage-wise effectiveness.}
We evaluate the individual contributions of temporal modeling at each stage by isolating the training of the sparse structure generator $\Phi_{\text{S}}$ and the \tslat generator $\Phi_{\text{L}}$. In \Cref{tab:ablation_main} (d), "w/o $\Phi_{\text{L}}$ training" denotes a configuration using our trained $\Phi_{\text{S}}$ alongside a frozen, frame-wise baseline generator $\mathcal{G}_{\text{L}}$~\cite{xiang2025structured}, whereas "w/o $\Phi_{\text{S}}$ training" represents the converse setup. Quantitatively, the "w/o $\Phi_{\text{L}}$ training" variant yields the highest per-frame appearance scores and competitive geometric metrics, while "w/o $\Phi_{\text{S}}$ training" degrades performance across nearly all metrics. However, qualitative analysis in \Cref{fig:appendix_ss_only} reveals that omitting $\Phi_{\text{L}}$ training causes severe temporal inconsistencies in texture, such as color drift and artifacts. These findings demonstrate that training both $\Phi_{\text{S}}$ and $\Phi_{\text{L}}$ is essential to preserve temporal consistency across geometry and appearance simultaneously.


\section{Analysis}
\label{sec:analysis}

\paragrapht{Attention visualization.}
\begin{figure}[t]
    \centering
    \includegraphics[width=\textwidth]{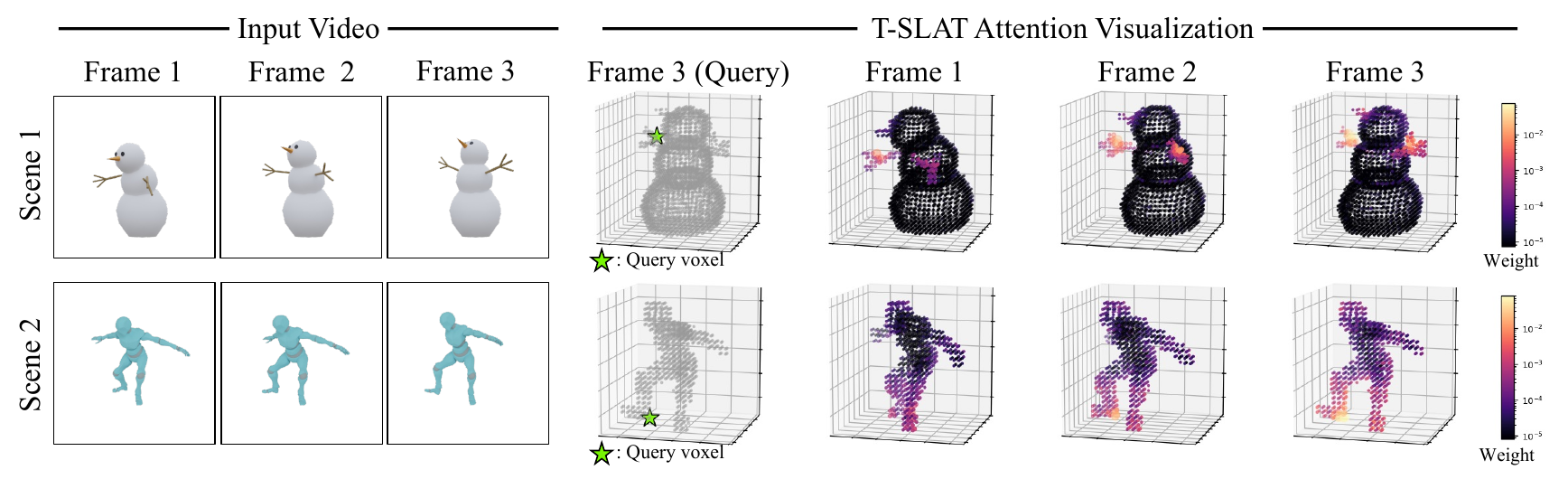}
    \caption{\textbf{Attention analysis in \tslat space.} For a query voxel (green star) in the target frame, we visualize its attention weights over voxel tokens of the conditioning frames. \ours\  captures geometric correspondences in the attention map and even exploits symmetry across frames, supporting temporally consistent voxel generation.}
    \label{fig:attention_analysis}
    \vspace{-10pt}
\end{figure}
To analyze how \ours\ encodes motion, we visualize the attention maps of our trained \tslat\ space. In \Cref{fig:attention_analysis}, we select a query token in the target voxel and visualize its attention weights over the voxel tokens in the previously generated (conditioning) voxels. In Scene 1, a query on the snowman's right hand attends to both hands in the previous frames, and in Scene 2, a query on the robot's right foot attends to both feet. These patterns indicate that \ours\ accurately establishes geometric correspondences in 3D space and even leverages symmetry cues across frames, producing temporally consistent voxel predictions.

\paragrapht{Error accumulation.}
\ours supports an autoregressive 4D generation while mitigating error accumulation. To verify that this design indeed mitigates the error accumulation in practice, we analyze how generation quality evolves as the video length grows. We compare against baselines~\cite{sabathier2026actionmesh,jiang2026mesh4d,zhang2025gaussian,ren2024l4gm}. As shown in~\Cref{fig:error_accum}, the quality of these baselines degrades progressively with longer videos, while \ours\ remains substantially more robust, maintaining stable geometry and appearance throughout the sequence. More analysis is in~\Cref{appendix:error_accum}.

\paragraph{Inference time.}






\begin{wraptable}{r}{0.65\textwidth} 
\vspace{-10pt}
\centering
\fontsize{7}{7}\selectfont
\setlength{\tabcolsep}{2pt}

\caption{\textbf{Inference time analysis.}}
\label{tab:inference_time}

\begin{tabular}{c|c|c|ccc|c|cc} 
\toprule
& 
& 
& \multicolumn{3}{c}{\textbf{Appearance}} 
& \multicolumn{1}{c}{\textbf{Video}} 
& \multicolumn{2}{c}{\textbf{Geometry}} \\
\cmidrule(lr){4-6} \cmidrule(lr){7-7} \cmidrule(lr){8-9} 
\textbf{Method} & \textbf{Steps} & \textbf{Time(s)} 
& LPIPS$\downarrow$ 
& CLIP$\uparrow$ 
& DreamSim$\downarrow$ 
& FVD$\downarrow$ 
& CD$\downarrow$ 
& F-score$\uparrow$ \\
\midrule
w/o Cache
& 25
& 111.42 
& 0.1591 & 0.8448 & {0.1987}
& 325.39 
& \textbf{0.1189} & {0.2084} \\

w/ Cache
& 25
& 54.90 
& \textbf{0.1576} & \textbf{0.8450} & \textbf{0.1966}
& \textbf{321.37}
& 0.1219 & \textbf{0.2088} \\

w/ Cache
& 12
& \textbf{28.03} 
& 0.1606 & 0.8399 & 0.2086 
& {326.58} 
& 0.1221 & 0.2029 \\

\bottomrule
\end{tabular}
\vspace{-9pt}
\end{wraptable}
\label{sec:inference_time}
We conduct an inference time analysis to evaluate the efficiency of \ours. With KV caching enabled, the average time required to process a 16-frame video on a single B200 GPU is reduced from 111.42 seconds to 54.90 seconds, corresponding to a $2.02\times$ speedup (See~\Cref{tab:inference_time}).
On top of it, the inference time can be reduced further by decreasing the number of denoising steps. When the denoising step is reduced from 25 to 12, time can be reduced to 28.03 seconds, a $3.89\times$ speedup. Detailed setups and analysis are reported in~\Cref{appendix:inference_time}.

\section{Conclusion}
We introduced \ours, an autoregressive 4D generative framework for unified dynamic 3D representations. By introducing \tslat, \ours jointly models temporal geometry and appearance while effectively handling complex motions and topological changes. We further proposed temporal-structural augmentation to enhance robustness and long-horizon consistency during autoregressive generation. Extensive experiments demonstrate that \ours achieves state-of-the-art performance in appearance and competitive results in geometry, while effectively mitigating error accumulation during long-horizon generation. We hope this work provides a significant step toward scalable and unified 4D generative modeling.

\newpage
\appendix
\section*{Appendix}
This appendix provides additional details and analyses of \ours. Section~\ref{appendix:implementation_details} details our training and dataset preparation pipeline. Section~\ref{sec:appendix_analysis} extends our error accumulation and inference time analysis. Section~\ref{sec:appendix_architecture} describes architectural details and our streaming inference procedure. Section~\ref{appendix:eval} elaborates on the geometry and appearance evaluation protocols. Section~\ref{sec:appendix_results} presents additional qualitative results, including novel view video generation and real-world generalization. Section~\ref{appendix:limitation} discusses limitations and social impact.

\section{Implementation Details}
\label{appendix:implementation_details}
\subsection{Training Details}
\label{appendix:training_details}
\begin{table}[h]
\centering
\caption{\textbf{Training configuration.}}
\label{tab:appendix_training_config}
\small
\setlength{\tabcolsep}{8pt}
\begin{tabular}{lcc}
\toprule
\textbf{Component} & \textbf{$\Phi_\text{S}$} & \textbf{$\Phi_\text{L}$} \\
\midrule
Optimizer               & AdamW                     & AdamW \\
Learning rate       & $1 \times 10^{-4}$      & $1 \times 10^{-4}$ \\
Optimizer betas         & $(0.9, 0.95)$          & $(0.9, 0.95)$ \\
Weight decay            & 0.0                      & 0.0 \\
Batch size              &  32                       & 64 \\
Window size             & $w=3$                    & $w=3$ \\
Objective                  & v-prediction & v-prediction \\
EMA decay               & 0.9999                   & 0.9999 \\
Gradient clipping       & 1.0                      & 1.0 \\
Trainable parameters & Self-attention \& AdaLN & Self-attention \& AdaLN \\
Condition image & $512 \times 512$ & $512 \times 512$ \\
Temporal augmentation & $k_t\sim\mathcal U(0,1)$ & $k_t\sim\mathcal U(0,1)$ \\
Structural augmentation & N/A & $\lambda=0.05$\\
\bottomrule
\end{tabular}
\end{table}

\subsection{Training Dataset}
\label{appendix:dataset}

In this section, we detail the data processing pipeline used to convert raw animated 3D assets into the encoded latents used for training. The pipeline consists of rigorous spatial alignment, multi-view and input video rendering, and quality filtering.

\paragrapht{Global normalization and temporal alignment.}
To prevent scale and translation jittering across time steps, we compute a union Axis-Aligned Bounding Box (AABB) evaluated across all sampled frames of a given sequence. The entire animated sequence is then globally translated and scaled using a single static center and scale factor. This ensures that the maximum motion envelope of the sequence is strictly confined within a canonical $[-0.5, 0.5]^3$ coordinate space, preserving the relative spatial scale of the character throughout the animation.

\paragrapht{Multi-view image rendering.}
To extract visual features from a pre-trained encoder~\cite{oquab2023dinov2}, we render 60 multi-view images per frame. Camera poses are uniformly distributed on a sphere using the Hammersley sequence with a random angular offset. The camera parameters are strictly fixed with a radius of $2.0$ and a Field of View (FoV) of $40^\circ$.

\paragrapht{Input video rendering.}
We render multi-view videos comprising 6 views per sequence. To ensure standard spatial reference, the primary view is strictly locked to the front. The remaining 5 views are distributed using the Hammersley sequence. Crucially, the cameras remain stationary across the temporal axis to provide stable video context. To prevent scaling artifacts across different views, we sample the inverse squared radius ($d = 1/r^2$) uniformly and dynamically derive the FoV within $[10^\circ, 70^\circ]$, so that the projected size of the unit bounding box remains constant across all conditioning cameras.

\paragrapht{Quality filtering.}
We implement a two-stage quality assurance protocol to exclude pathological meshes that destabilize training. (i) Topology filtering: we compute the bounding box extents for all individual sub-meshes within a scene. If the maximum sub-mesh extent exceeds the median sub-mesh extent by a factor greater than $15.0$, the entire sequence is discarded. This effectively filters out models containing disconnected "floaters" or extreme artifact spikes that would otherwise crush the main character into a fraction of the voxel grid. (ii) Voxel density filtering: all meshes are discretized into a $64^3$ voxel grid. We discard the sample if any single frame contains fewer than $500$ occupied voxels. This prevents the inclusion of degraded animations where the mesh topology collapses or moves outside the normalized bounds during motion.

\section{Further Analysis}
\label{sec:appendix_analysis}
In~\Cref{sec:analysis}, we analyze error accumulation and inference time of \ours. In~\Cref{sec:appendix_analysis}, we further demonstrate that \ours\ exhibits minimal error accumulation compared to baselines while maintaining short inference time. 

\subsection{Error Accumulation}
\label{appendix:error_accum}
\begin{figure}[h]
    \centering

    \begin{subfigure}[t]{0.48\linewidth}
        \centering
        \includegraphics[width=\linewidth]{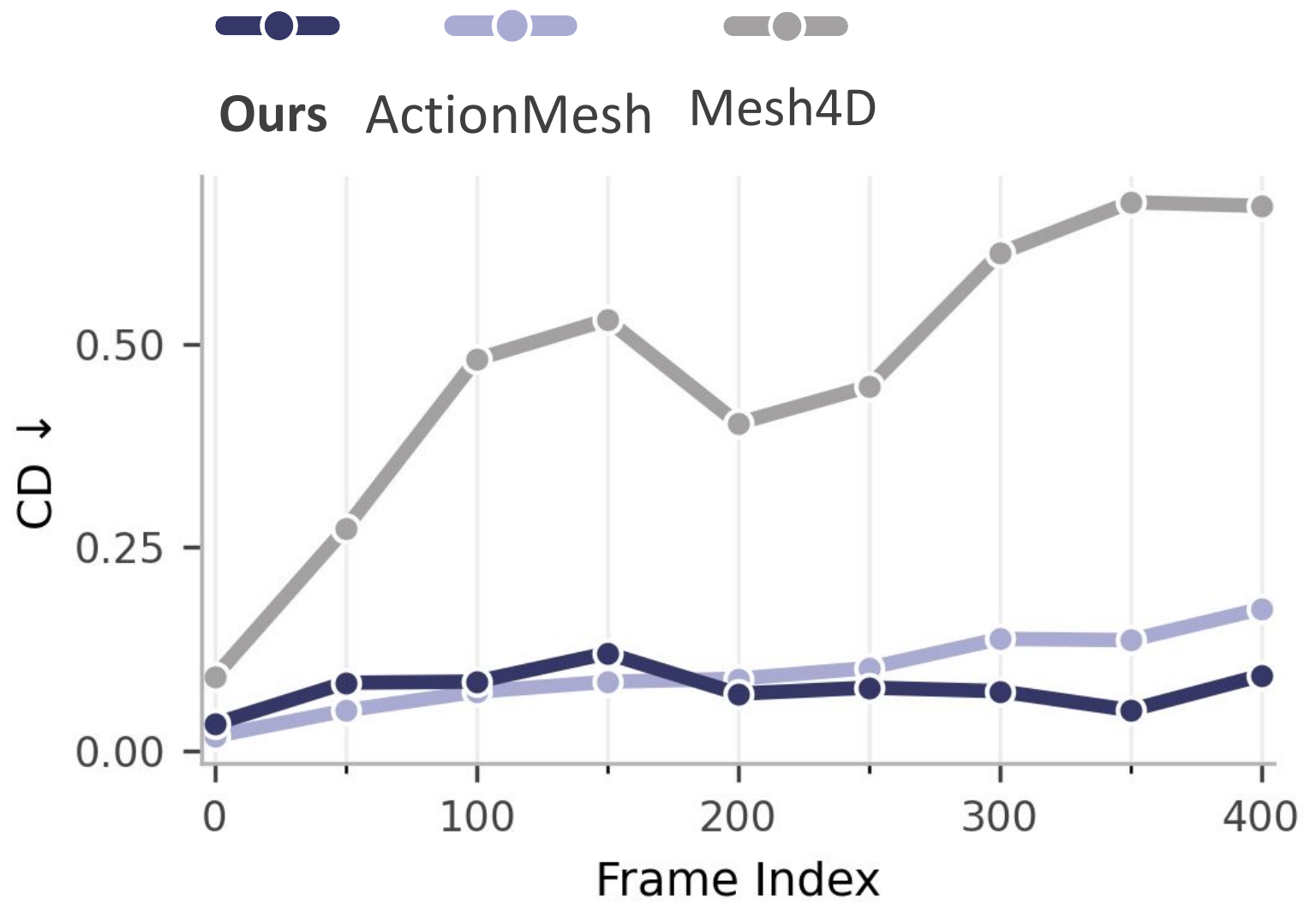}
        \caption{CD}
    \end{subfigure}
    \hfill
    \begin{subfigure}[t]{0.48\linewidth}
        \centering
        \includegraphics[width=\linewidth]{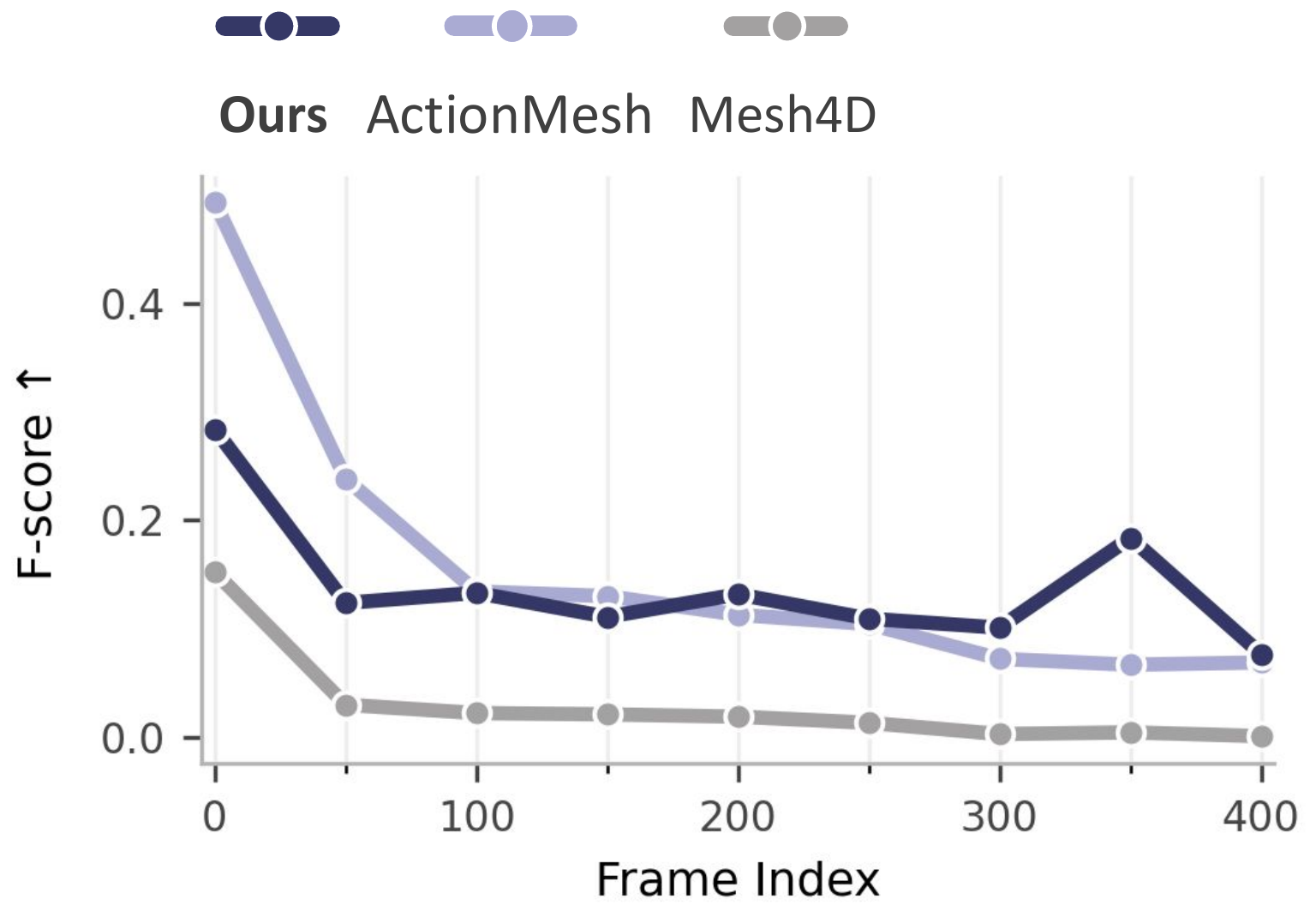}
        \caption{F-score}
    \end{subfigure}

    \vspace{5pt}

    \begin{subfigure}[t]{0.48\linewidth}
        \centering
        \includegraphics[width=\linewidth]{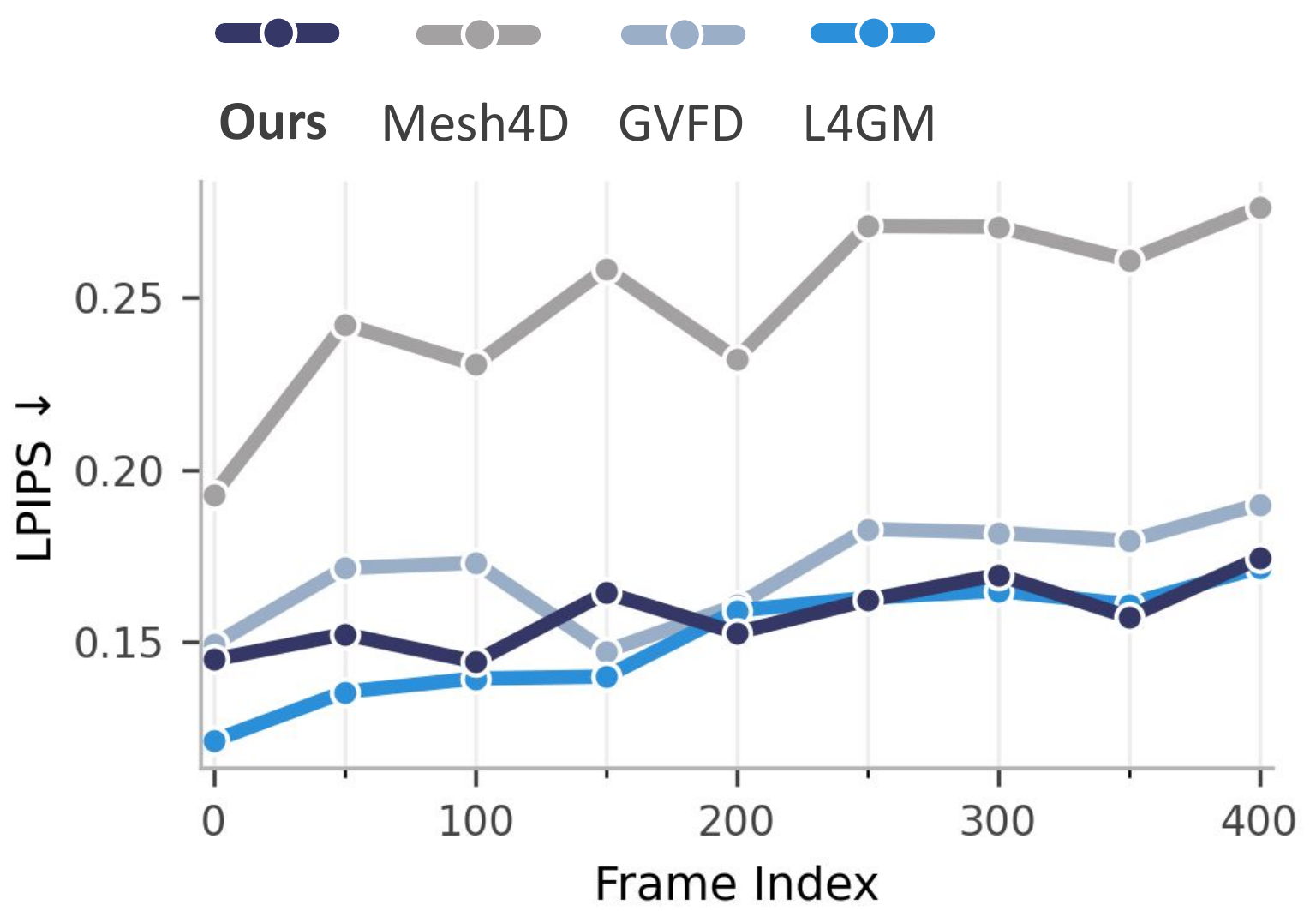}
        \caption{LPIPS}
    \end{subfigure}
    \hfill
    \begin{subfigure}[t]{0.48\linewidth}
        \centering
        \includegraphics[width=\linewidth]{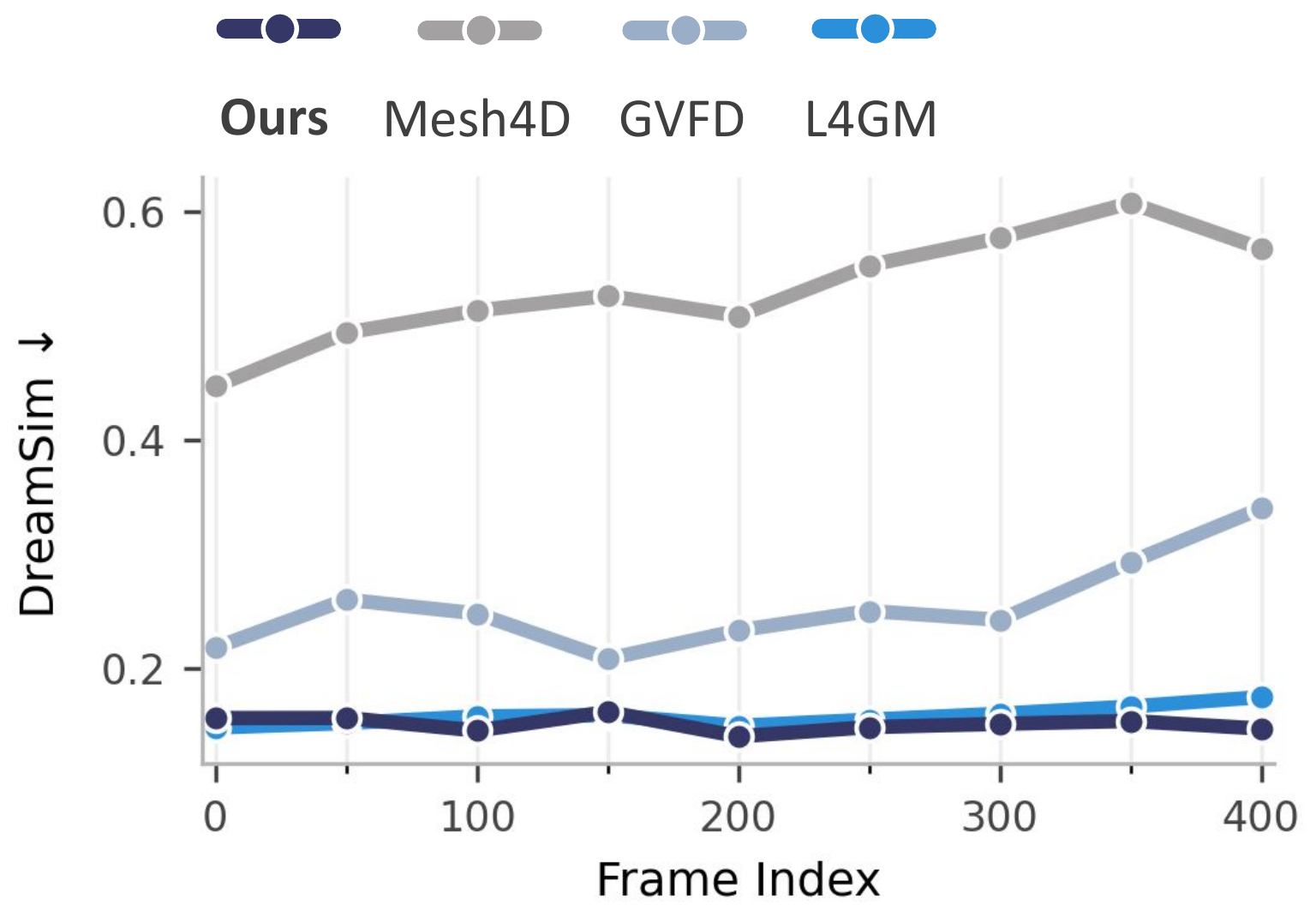} 
        \caption{DreamSim}
    \end{subfigure}

\caption{\textbf{Long-horizon error accumulation analysis.} We compare the temporal evolution of geometric and appearance metrics across frames. Our method maintains consistently stable performance over time with minimal degradation, demonstrating strong robustness against long-term error accumulation compared to prior methods.}
    \label{fig:appendix_error}
\end{figure}
We study the effect of video length on generation quality to analyze error accumulation in long video generation on Motion80-long~\cite{chen2026motion}. Specifically, we evaluate how performance changes as the frame length increases. 

\paragrapht{Geometry.}
In~\Cref{fig:appendix_error}(a) and (b), Mesh4D~\cite{jiang2026mesh4d} and ActionMesh~\cite{sabathier2026actionmesh} exhibit a noticeable increase in Chamfer Distance and a corresponding decrease in F-score as the frame index grows. This signifies accumulated prediction errors that progressively distort geometry over long videos. \ours, in contrast, maintains stable geometry throughout the sequence.

\paragrapht{Appearance.}
In~\Cref{fig:appendix_error}(c) and (d), Mesh4D~\cite{jiang2026mesh4d} and GVFD~\cite{zhang2025gaussian} exhibit similar degradation as the sequence becomes longer, due to accumulated prediction errors over time. \ours\ maintains stable performance throughout the sequence, indicating that our autoregressive formulation in the \tslat\ space effectively mitigates temporal drift and enables robust long-horizon generation.

\subsection{Inference Time}
\begin{wraptable}{r}{0.6\textwidth}
\vspace{-10pt}
\centering
\fontsize{6}{7}\selectfont
\setlength{\tabcolsep}{2pt}

\caption{\textbf{Inference time analysis.} The shortcut (SC) finetuning accelerates SS inference by reducing the number of denoising steps while maintaining comparable performance. Columns \textbf{$\Phi_\text{S}$} and \textbf{$\Phi_\text{L}$} indicate the number of denoising steps.}
\label{tab:app_inference_time}

\begin{tabular}{c|c|c|c|ccc|c|cc}
\toprule
& & & 
& \multicolumn{3}{c}{\textbf{Appearance}} 
& \multicolumn{1}{c}{\textbf{Video}}
& \multicolumn{2}{c}{\textbf{Geometry}} \\
\cmidrule(lr){5-7} \cmidrule(lr){8-8} \cmidrule(lr){9-10}
\textbf{Caching}
& \textbf{$\Phi_\text{S}$}
& \textbf{$\Phi_\text{L}$}
& \textbf{Time(s)} 
& LPIPS$\downarrow$ 
& CLIP$\uparrow$ 
& DreamSim$\downarrow$ 
& FVD$\downarrow$
& CD$\downarrow$ 
& F-score$\uparrow$ \\
\midrule

\checkmark & 25 & 25
& 54.90 & 0.1904 & \textbf{0.8551} & \textbf{0.1857} & 
203.02 & \textbf{0.0972} & \textbf{0.2138} \\


\checkmark & 4(SC) & 12
& 20.71
& \textbf{0.1873} & 0.8500 & 0.1925 & \textbf{192.05}
& 0.1055 & 0.2028 \\

\checkmark & 2(SC) & 12
& 20.52
& 0.1902 & 0.8423 & 0.2030 & 221.71
& 0.1067 & 0.1913 \\

\bottomrule
\end{tabular}
\vspace{-9pt}
\end{wraptable}
\label{appendix:inference_time}

\paragraph{Analysis setup.}
The runtime is measured on ActionBench~\cite{sabathier2026actionmesh} by generating 16 frames, averaged over 128 scenes, on a single NVIDIA B200 GPU.

\paragraph{Shortcut model.}
The inference time can be further reduced with shortcut fine-tuning~\cite{frans2024one}. Following SAM3D~\cite{chen2025sam}, we apply the shortcut finetuning to the sparse structure stage $\Phi_\text{S}$. This enables direct approximation of few-step denoising trajectories, thereby accelerating the generation by reducing the number of denoising steps.
\begin{figure}[h]
    \centering
    \includegraphics[width=\linewidth]{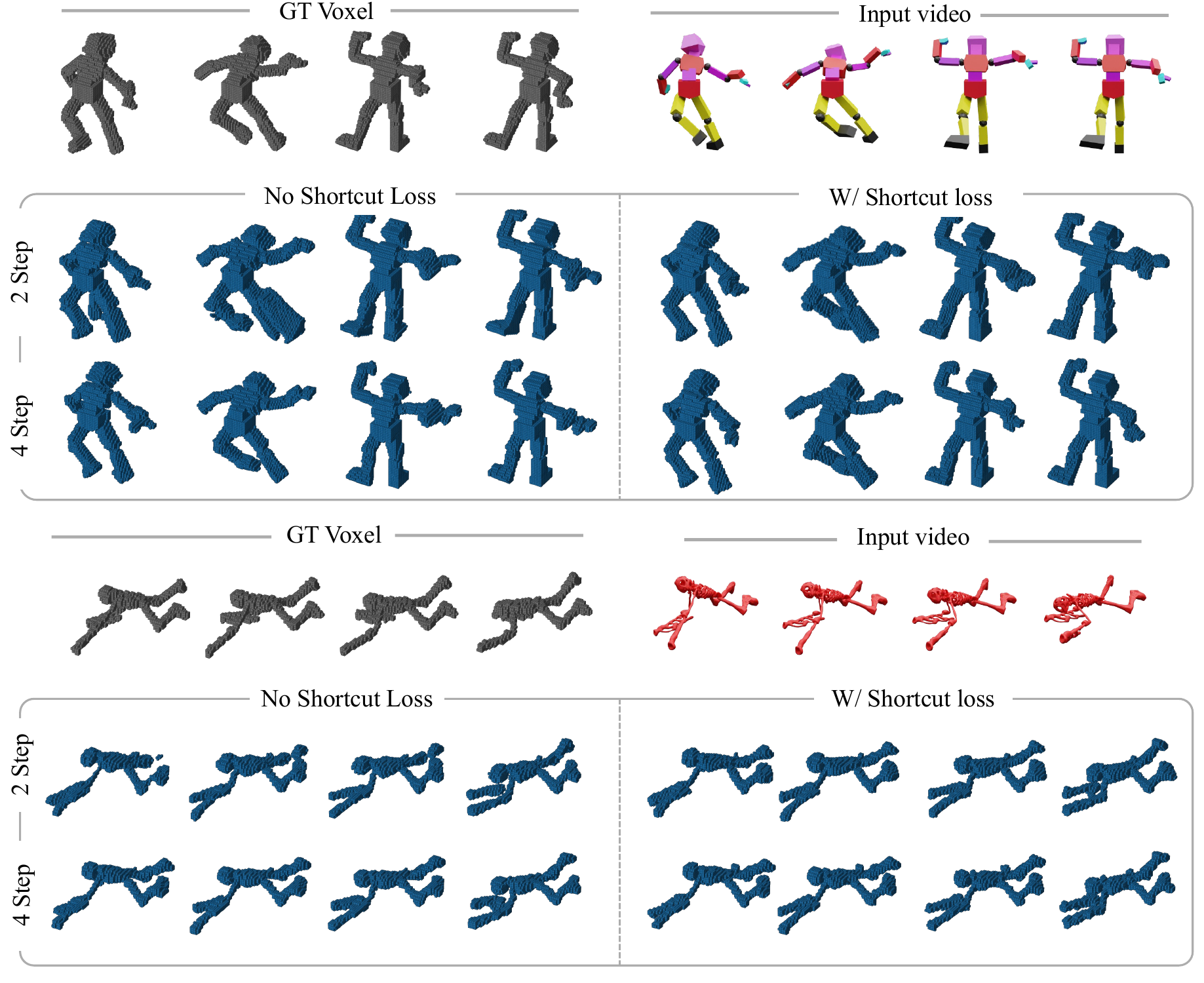}
    \caption{\textbf{SS stage with Shortcut fine-tuning.} Without shortcut fine-tuning, the model produces noticeable artifacts. In contrast, applying shortcut loss fine-tuning significantly reduces these artifacts and improves output quality.}
    \label{fig:appendix_few_step}
\end{figure}

We evaluate the effect of reducing the number of denoising steps for the $\Phi_\text{S}$ stage using the proposed shortcut model. \Cref{fig:appendix_few_step} shows the effect of shortcut finetuning qualitatively. Without the shortcut fine-tuning, we observe noticeable structural artifacts, where object parts such as legs or arms are often disconnected or incorrectly assembled. In contrast, applying the finetuning improves structural consistency and reduces such artifacts, leading to more coherent geometry.

\Cref{tab:app_inference_time} summarizes the effect of reducing the denoising timesteps on inference efficiency. With 25 denoising steps for both SS and \slat generation, the model takes 54.90 seconds. Applying shortcut inference to the $\Phi_\text{S}$ stage enables extreme 4-step and 2-step denoising, resulting in substantial acceleration to 20.71 seconds. Although aggressive timestep reduction introduces a moderate degradation in metrics, the results demonstrate a favorable trade-off between inference efficiency and generation quality. Overall, shortcut fine-tuning reduces inference cost while maintaining competitive appearance and geometric fidelity.

\section{Architecture Details}
\label{sec:appendix_architecture}
To adapt the 3D generation capabilities of the original \slat framework~\cite{xiang2025structured} to the temporal domain, we modify the core flow transformer architecture. The original \slat consists of two flow transformers, $\Phi_\text{S}$ (sparse structure generation) and $\Phi_\text{L}$ (\tslat generation), both utilizing a transformer backbone with 24 blocks. In the original architecture, each block follows a sequence of self-attention, cross-attention, and a Feed-Forward Network (FFN). The noise level $k_t$ is injected via Adaptive Layer Normalization (AdaLN). For visual conditioning, DINOv2~\cite{oquab2023dinov2} features are extracted from the input video and integrated through cross-attention layers in each block.

We introduce the following modifications for \ours:
\begin{itemize}
    \item \textbf{Causal attention:} We inflate the self-attention to causal 3D self-attention layers by concatenating each frame tokens into a single sequence. This ensures that tokens from frame $t$ attend only to themselves and tokens from preceding frames $t' \leq t$, maintaining the autoregressive property.
    \item \textbf{Noise Sampling:} While the original framework utilizes a Logit-Normal noise schedule \cite{xiang2025structured}, we adopt a uniform sampling strategy for the per-frame noise level $k_t \sim \mathcal{U}(0, 1)$ during training. Exposing the model to a wider and more diverse range of noise scales across the temporal sequence, we encourage the models to learn robust denoising capabilities under varying levels of historical artifacts.
    \item \textbf{Parameter Efficiency:} During finetuning, we only fine-tune the 3D self-attention and AdaLN parameters and keep other weights frozen to preserve the robust 3D generative priors learned from TRELLIS~\cite{xiang2025structured}.
\end{itemize}

\subsection{VAE and decoders}
\textsc{Morphos} leverages the existing high-fidelity encoding and decoding pipeline of the base 3D model:
\begin{itemize}
    \item \textbf{Independent Encoding:} Each frame of the input video is encoded into the structured latent space independently using the pre-trained TRELLIS encoders.
    \item \textbf{Shared Decoders:} We utilize the original decoders for meshes, 3D Gaussians, and Radiance Fields without modification.
    \item \textbf{Consistency via Flow:} We find that maintaining frozen encoders and decoders, focusing solely on temporal modeling within the flow transformers, is sufficient to achieve temporal consistency.
\end{itemize}

\subsection{Inference details}
Our streaming inference enables the generation of arbitrary-length sequences through an efficient autoregressive process. The generation begins with the first frame ($t=1$). Since there is no prior history, the flow transformer performs denoising by attending solely to the noisy latent of the current frame. Once the latent $\mathbf{z}^1$ is fully denoised (referred to as the ``clean'' latent), it is passed through the transformer blocks one final time. During this forward pass, we compute and store the Key and Value tensors for all tokens of the frame in a local cache. For subsequent frames ($t \geq 2$), instead of re-processing the entire history, the model simply retrieves these pre-computed KV pairs from the cache. This allows the current query tokens to attend to historical context with minimal computational overhead.

To maintain a constant memory footprint, we implement a fixed-size cache management strategy with a cache size of 2 (as we set the window size $w=3$). When the number of stored frames reaches this capacity, an eviction policy is triggered. This approach ensures that while the model always has access to the most immediate and relevant temporal context, the memory requirements do not scale linearly with the total sequence length, enabling the generation of very long 4D animations on a single GPU.

\section{Evaluation Details}
\label{appendix:eval}


We evaluate all models on three datasets: Motion80~\cite{chen2026motion}, ActionBench~\cite{sabathier2026actionmesh}, and Consist4D~\cite{jiang2023consistent4d}. Each dataset provides an input video $\mathbf{I}^{1:T} = \{\mathbf{I}^t\}_{t=1}^{T}$, which is fed to all models to obtain predicted meshes $\hat{\mathcal{O}}^{1:T} = \{\hat{\mathcal{O}}^t\}_{t=1}^{T}$, where $\hat{\mathcal{O}}^t=\langle \hat{\mathcal{V}}^t, \hat{\mathcal{F}}^t, \hat{\mathcal{T}}^t\rangle$. The datasets, however, differ in the ground-truth supervision they provide for evaluation. Motion80~\cite{chen2026motion} and ActionBench~\cite{sabathier2026actionmesh} provide ground-truth mesh sequences $\mathcal{O}^{1:T} = \{\mathcal{O}^t\}_{t=1}^T$, where $\mathcal{O}^t=\langle \mathcal{V}^t, \mathcal{F}^t, \mathcal{T}^t\rangle$ denotes the mesh of frame $t$, but do not include rendered evaluation videos. Consist4D~\cite{jiang2023consistent4d}, in contrast, provides evaluation ground-truth videos but no ground-truth meshes.    To enable a unified appearance evaluation protocol across all three datasets, we render the ground-truth meshes from Motion80 and ActionBench into evaluation videos. Specifically, we render $\mathcal{O}^{1:T}$ from four canonical viewpoints, $\theta \in \{0^\circ, 90^\circ, 180^\circ, 270^\circ\}$, yielding evaluation videos $\{\mathbf{I}^t_{\theta}\}_{t=1}^T$ at a resolution of $256 \times 256$ with RGBA channels. For Consist4D, we directly use the provided evaluation videos. This unified setup allows us to evaluate all models under a consistent protocol across datasets.

\subsection{Appearance Evaluation}
\label{appendix:appearance_eval}



For appearance evaluation, we render $\hat{\mathcal O}^{1:T}$ to generate predicted video $\{\mathbf{\hat{I}}^t_{\theta}\}_{t=1}^T$ and 
conduct azimuth alignment to ensure accurate comparison across methods. 
Specifically, we render $\hat{\mathbf{I}}^t_{\theta}$ over a predefined azimuth set $\Theta$, and select the reference view by minimizing the $\ell_2$ distance to the evaluation ground-truth frame:
\[
\theta^* = \arg\min_{\theta \in \Theta} \left\| \mathbf{I}^1_{0^\circ} - \hat{\mathbf{I}}^1_{\theta} \right\|_2.
\]
 
For Motion80~\cite{chen2026motion} and ActionBench~\cite{sabathier2026actionmesh}, we set $\Theta = \{0^\circ, 90^\circ, 180^\circ, 270^\circ\}$. The selected view $\theta^*$ is used as the reference, and the remaining views are rendered accordingly. For Consist4D~\cite{jiang2023consistent4d}, we instead follow the original camera setup of Consist4D~\cite{jiang2023consistent4d}.

After determining $\theta^*$, we treat it as $0^\circ$ and render the predicted mesh to generate videos $\{\hat{\mathbf{I}}^t_{\theta}\}_{t=1}^T$ where $\theta \in \{0^\circ, 90^\circ, 180^\circ, 270^\circ\}$. This ensures that the predicted videos are rendered under the same viewpoint configuration as the ground truth, with a white background at a resolution of $256 \times 256$. Using this aligned viewpoint, we compute evaluation metrics LPIPS, CLIP, and DreamSim for each of the four views independently and calculate the final results by averaging across all views. For video-level evaluation, we measure Fréchet Video Distance (FVD)~\cite{unterthiner2019fvd}. 


\subsection{Geometry Evaluation}
\label{appendix:geometric_eval}

For geometry evaluation of meshes, we measure temporal and per-frame reconstruction quality using Chamfer Distance (CD), F-score, and Point-to-Surface distance (P2S). 
CD measures temporal consistency by first aligning the predicted and ground-truth mesh sequences using the Iterative Closest Point (ICP) algorithm~\cite{besl1992method} on the first frame, and then averaging the Chamfer Distance across the sequence. For F-score and P2S, we use the same global alignment as CD to ensure temporally consistent evaluation. 

To ensure a comprehensive geometric evaluation, we adopt a set of complementary metrics from prior works. Specifically, we use the F-score from Motion324~\cite{chen2026motion}, Chamfer Distance (CD) from ActionMesh~\cite{sabathier2026actionmesh}, and Point-to-Surface (P2S) from Mesh4D~\cite{jiang2026mesh4d}. These metrics are combined to provide a balanced assessment of geometric accuracy, temporal consistency, and surface fidelity.

For accurate geometric evaluation, we first normalize the ground-truth vertices $\mathcal{V}^t$ into a canonical space of $[-1, 1]^3$. For each frame $t$, we uniformly sample $P = 100{,}000$ points from $\mathcal{V}^t$ and $\hat{\mathcal{V}}^t$, which we denote as
\begin{equation}
\mathcal{W}^t = \{\mathbf{w}^t_i\}_{i=1}^{P}, \quad
\hat{\mathcal{W}}^t = \{\hat{\mathbf{w}}^t_i\}_{i=1}^{P},
\end{equation}
respectively. These sampled point sets serve as the basis for all geometric metrics (CD, F-score, P2S).
 
To estimate the alignment transformation $\zeta_0$, we additionally sample 3{,}000 points from the vertex sets $\mathcal{V}^1$ and $\hat{\mathcal{V}}^1$ of the first frame and run the Iterative Closest Point (ICP) algorithm~\cite{besl1992method}. The resulting $\zeta_0$ is then applied consistently to $\hat{\mathcal{W}}^t$ for all $t$, enabling evaluation under a shared global alignment.

Given two point sets $\mathcal{W}_1$ and $\mathcal{W}_2$ each containing $P$ points, the Chamfer Distance is defined as
\begin{equation}
\mathrm{CD}(\mathcal{W}_1, \mathcal{W}_2) =
\frac{1}{P} \sum_{\mathbf{u} \in \mathcal{W}_1} \min_{\mathbf{v} \in \mathcal{W}_2} \|\mathbf{u} - \mathbf{v}\|_2^2
+
\frac{1}{P} \sum_{\mathbf{v} \in \mathcal{W}_2} \min_{\mathbf{u} \in \mathcal{W}_1} \|\mathbf{u} - \mathbf{v}\|_2^2.
\end{equation}

We compute $\mathrm{CD}(\mathcal{W}^t, \zeta_0(\hat{\mathcal{W}}^t))$ for all $t$ using the single alignment $\zeta_0$ estimated from the initial frame, which enables evaluation under a shared global alignment. The final CD score is averaged over the sequence:
\begin{equation}
\mathrm{CD} = \frac{1}{T} \sum_{t=1}^{T} \mathrm{CD}\big(\mathcal{W}^t, \zeta_0(\hat{\mathcal{W}}^t)\big).
\end{equation}









We evaluate the F-score at a threshold $\tau = 0.01$. Precision measures the fraction of predicted points that lie within $\tau$ of the ground-truth surface, and recall measures the reverse:
\begin{equation}
\mathrm{Precision} = \frac{1}{T\times P} \sum_{t=1}^{T} \sum_{i=1}^{P}
\mathbf{1}\!\left( \min_{j \in \{1,\dots,P\}} \big\| \zeta_0(\hat{\mathbf{w}}^t_i) - \mathbf{w}^t_j \big\|_2 < \tau \right),
\end{equation}
\begin{equation}
\mathrm{Recall} = \frac{1}{T \times P} \sum_{t=1}^{T} \sum_{i=1}^{P}
\mathbf{1}\!\left( \min_{j \in \{1,\dots,P\}} \big\| \mathbf{w}^t_i - \zeta_0(\hat{\mathbf{w}}^t_j) \big\|_2 < \tau \right),
\end{equation}
where $\mathbf{1}(\cdot)$ denotes the indicator function, $i$ indexes the source point set, and $j$ indexes the target point set being searched over. The F-score is defined as the harmonic mean of precision and recall:
\begin{equation}
\text{F-score} =
\frac{2 \cdot \mathrm{Precision} \cdot \mathrm{Recall}}
{\mathrm{Precision} + \mathrm{Recall}}.
\end{equation}

We further evaluate the Point-to-Surface (P2S) distance to measure the discrepancy between predicted and ground-truth mesh surfaces. Unlike Chamfer Distance, which relies on point-to-point comparisons and is sensitive to sampling density, P2S computes the shortest Euclidean distance from sampled points to the continuous surface of the target mesh, providing a more robust assessment of surface accuracy. Using the same sampled point sets $\mathcal{W}^t$ and $\hat{\mathcal{W}}^t$, for each frame we compute the forward distance from $\zeta_0(\hat{\mathcal{W}}^t)$ to $\mathcal{F}^t$ and the backward distance from $\mathcal{W}^t$ to $\zeta_0(\hat{\mathcal{F}}^t)$; the per-frame P2S is defined as their average, and the final score is obtained by averaging over all frames.

\subsection{Baseline inference}
\label{appendix:inference_details}

When the input video length exceeds the model's input size, we inference chunk-wise auto-regressively to handle long sequences. Specifically, the input sequence is divided into consecutive chunks, and the last-frame mesh from the previous chunk is used as the initial mesh condition for the next chunk, enabling temporally consistent predictions. However, for Gaussian-based models such as L4GM~\cite{ren2024l4gm} and GVFD~\cite{zhang2025gaussian}, which do not take mesh inputs, each chunk is processed independently without initial mesh conditioning. In contrast, when the input video length is shorter than the model's input size, we pad the input by repeating the last frame until it matches the model's input size, and then perform inference on the padded sequence.


\section{Additional Results}
\label{sec:appendix_results}

\paragrapht{Additional qualitative results.}
\begin{figure}[t]
    \centering
    \includegraphics[width=\textwidth]{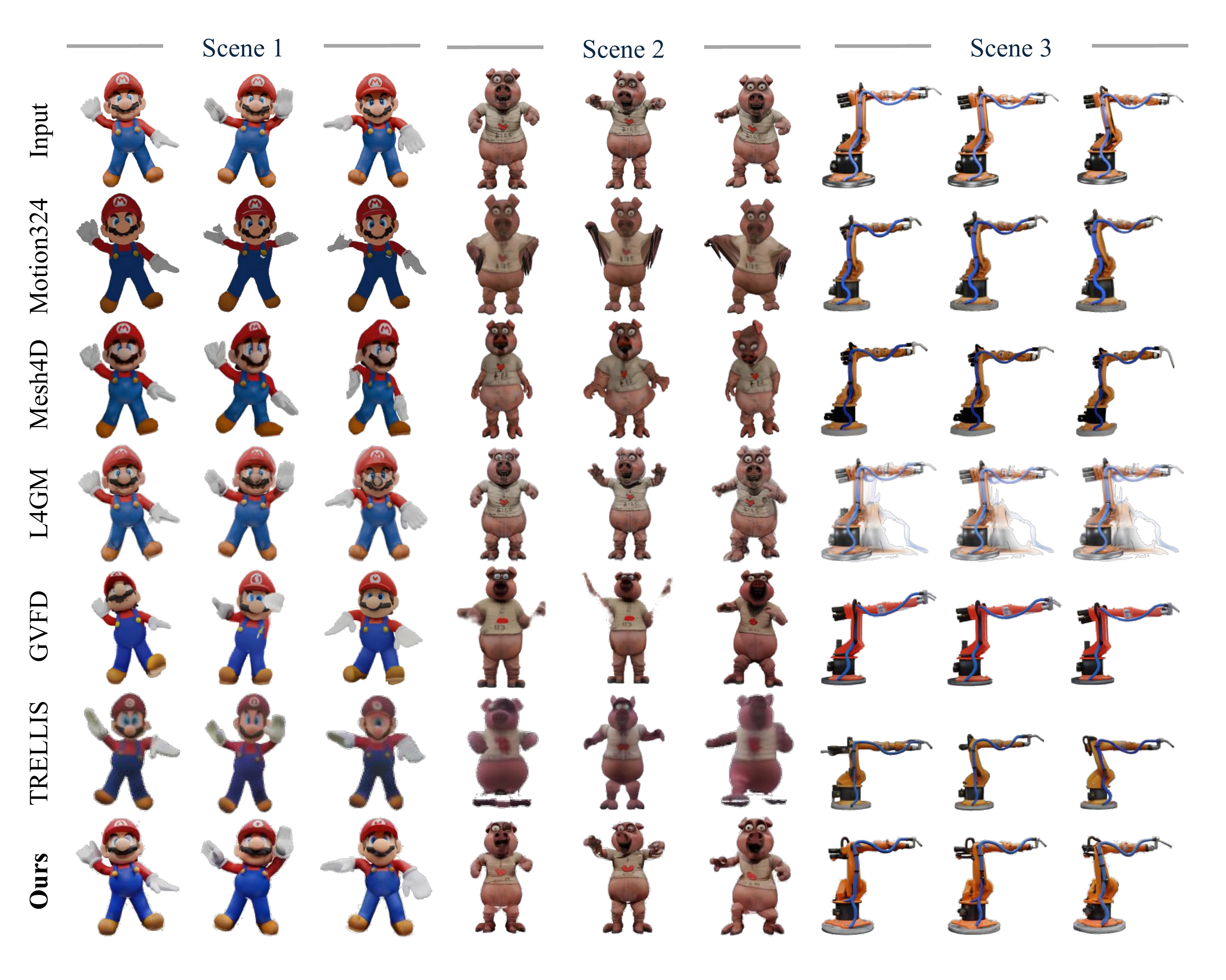}
    \caption{\textbf{Additional qualitative results on appearance.} We compare rendered videos of \ours\ against baselines~\cite{chen2026motion,jiang2026mesh4d,zhang2025gaussian,ren2024l4gm,xiang2025structured} on additional input sequences. \ours\ preserves fine-grained texture details and maintains temporally stable appearance across frames.} 
    \label{fig:qual_appearance_2}
    \vspace{-10pt}
\end{figure}
\begin{figure}[t]
    \centering
    \includegraphics[width=\textwidth]{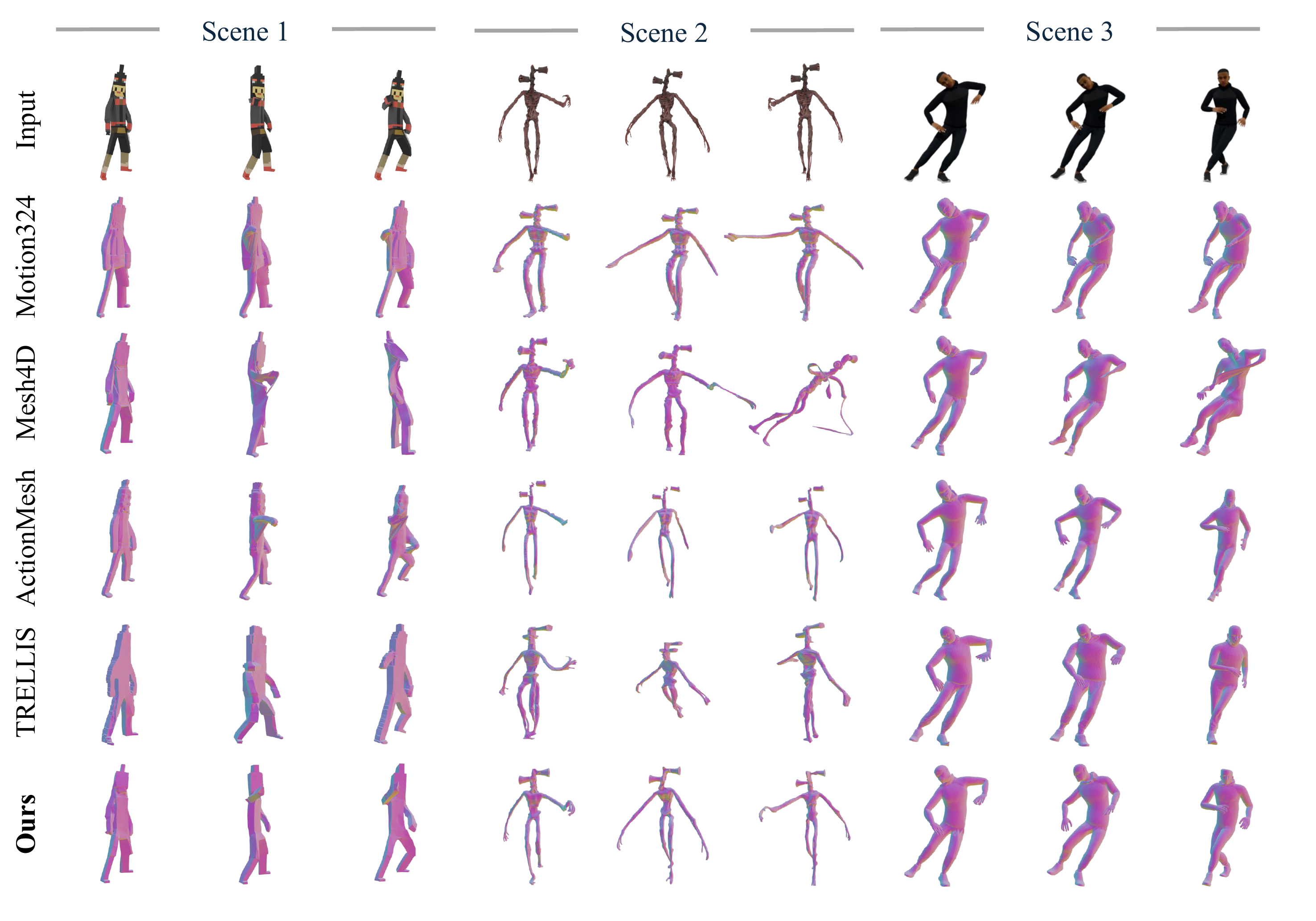}
    \vspace{-10pt}
    \caption{\textbf{Additional qualitative results on geometry.} Rendered normal maps of generated mesh sequences from \ours\ and baselines~\cite{chen2026motion,jiang2026mesh4d,sabathier2026actionmesh,xiang2025structured}. \ours\ generates meshes with sharp surface details and consistent geometry throughout the sequence.}
    \label{fig:qual_geo_2}
    \vspace{-11pt}
\end{figure}
We provide additional qualitative comparisons between \ours\ and the baselines across a broader set of sequences and object categories. \Cref{fig:qual_appearance_2} shows appearance comparisons~\cite{chen2026motion,jiang2026mesh4d,zhang2025gaussian,ren2024l4gm,xiang2025structured}; consistent with the main paper, \ours\ produces temporally consistent renderings
over time. \Cref{fig:qual_geo_2} shows the corresponding normal maps ~\cite{chen2026motion,jiang2026mesh4d,sabathier2026actionmesh,xiang2025structured}, where \ours\ recovers coherent surface geometry with well-defined structural details and stable topology across frames. These additional examples confirm that both the appearance and geometric improvements hold across diverse cases rather than being specific to the sequences shown in the main paper.

\paragrapht{Novel view video generation.} 
\begin{figure}[t]
    \centering
    \includegraphics[width=\textwidth]{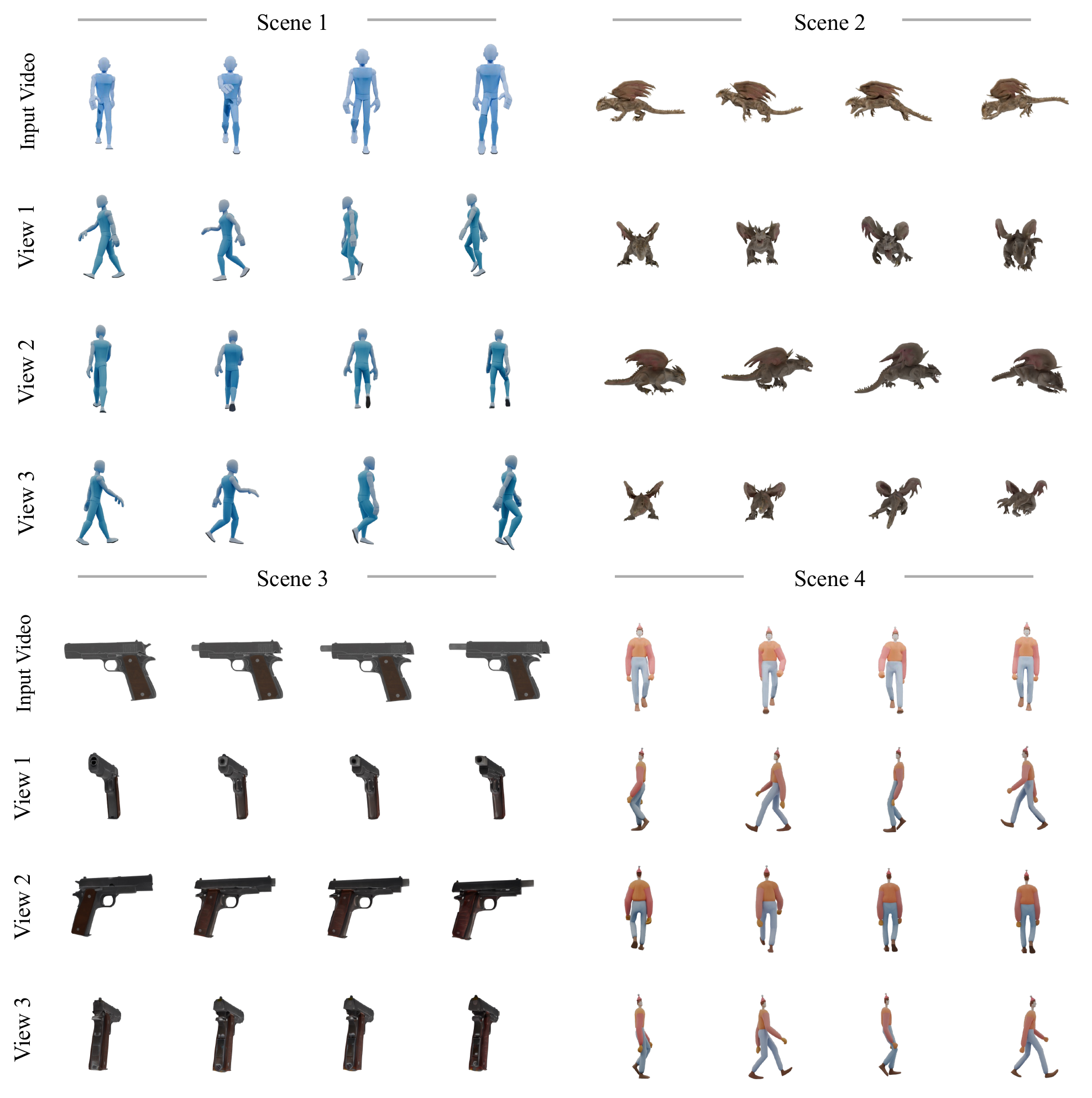}
    \caption{\textbf{Novel view video generation.} Given an input video, \ours\ reconstructs a dynamic 4D representation, enabling high-quality rendering from novel viewpoints with consistent geometry and appearance over time.}
    \label{fig:appendix_qual_appearance}
\end{figure}
Given an input video, our model reconstructs a dynamic 4D representation, which enables rendering from arbitrary viewpoints. \Cref{fig:appendix_qual_appearance} shows novel view video generation results. 
The rendered videos exhibit strong temporal consistency and stable appearance across frames, while preserving fine-grained details under viewpoint changes. \ours maintains coherent geometry and texture over time, resulting in visually consistent novel view renderings.

\begin{figure}[t]
    \centering
    \includegraphics[width=\textwidth]{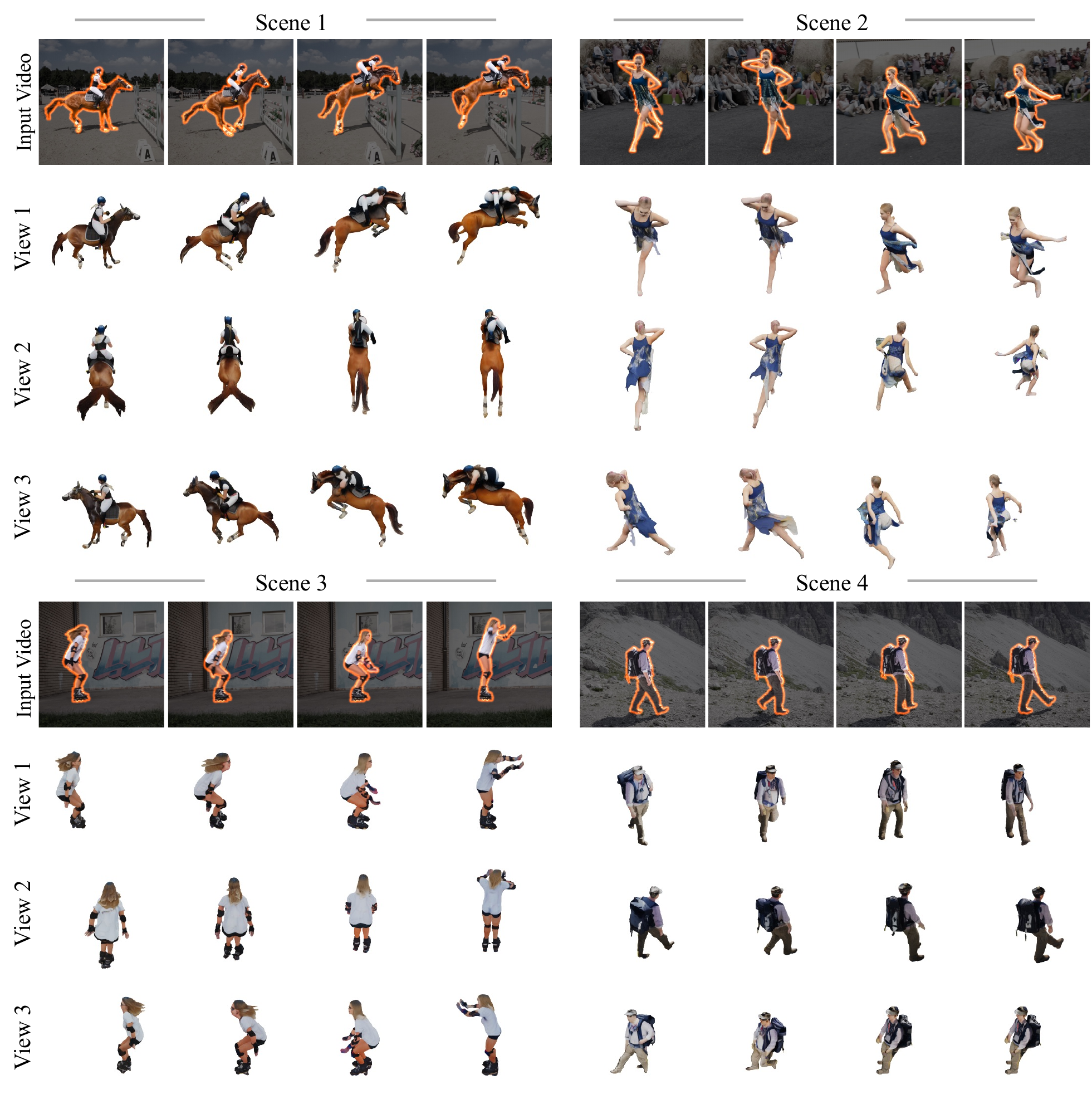}
    \vspace{-10pt}
    \caption{\textbf{Generalization to real-world domain~\cite{pont20172017}.} Given real-world videos from DAVIS, \ours\ generates temporally consistent 4D results with stable geometry and coherent appearance. Our method further enables novel-view video generation.}
    \label{fig:appendix_davis}
    \vspace{-10pt}
\end{figure}

\paragrapht{Generalization to real domain.}
To demonstrate that \ours\ generalizes beyond synthetic settings, we perform inference on real-world videos from the DAVIS dataset~\cite{pont20172017}. Given a real-world video, we use SAM2~\cite{ravi2024sam} to obtain object masks, which are then used as inputs to our model. As shown in~\Cref{fig:appendix_davis}, \ours\ produces temporally consistent results with stable geometry and coherent appearance, despite the increased complexity and noise of real-world inputs. Moreover, our model enables novel-view video generation from these real-world inputs, demonstrating its ability to reconstruct and render consistent 4D dynamics across viewpoints. These results indicate that \ours\ effectively transfers to real-world scenarios.

\section{Limitations \& Social Impact}
\label{appendix:limitation}

\paragrapht{Limitations.}
While \ours\ demonstrates strong performance across diverse 4D generation settings, several limitations remain. First, although our approach introduces a unified latent representation for multiple 3D formats, the decoding quality may still vary depending on the target representation. Second, the current model relies on a fixed set of training assets and motion patterns, which may limit generalization to rare or out-of-distribution motion dynamics. Expanding the diversity and scale of training data could further improve robustness and coverage.

\paragrapht{Social impact.}
The proposed framework contributes to the advancement of 4D content generation, enabling more flexible and efficient creation of dynamic 3D assets from video inputs. This has potential applications in content creation, virtual reality, gaming, and simulation, where rapid generation of high-quality 3D assets can significantly reduce production costs and time. At the same time, as with other generative technologies, there are potential risks associated with misuse, including the generation of misleading or non-consensual digital content.

\clearpage
{\small
\bibliographystyle{plain}
\bibliography{references}
}


\end{document}